\newcommand{\Bem}[1]{ }

\documentclass{article}

\title{Explainable Graph Spectral Clustering For GloVe-like Text Embeddings}
\newcommand{\orcidID}[1]{$^{#1}$ }
\author{Mieczysław A. Kłopotek \orcidID{0000-0003-4685-7045}
        \and
        Sławomir T. Wierzchoń \orcidID{0000-0001-8860-392X} \and Bartłomiej Starosta \orcidID{0000-0002-5554-4596}  \and Piotr Borkowski \orcidID{0000-0001-9188-5147} 
        \and  Dariusz Czerski \orcidID{0000-0002-3013-3483} 
        \and Eryk Laskowski 
        \orcidID{0000-0001-6346-6324}
        \\ 
        Institute of Computer Science of Polish Academy of Sciences\\ 
  ul. Jana Kazimierza 5, 01-248 Warszawa, Poland\\
        }

\usepackage{graphicx} 
\usepackage{url}
\usepackage[T1]{fontenc}
\usepackage{hyperref} 
\usepackage{amsmath} 
\usepackage{amsfonts}
\usepackage{placeins}

\newcommand{\Vol}{\mathcal{V}}
\newcommand{\dok}{\delta}
\newcommand{\DokSet}{\mathcal{D}}

\newcommand{\tht}[1]{\texttt{\##1}}
\newcommand{\vct}[2]{\parbox[t][#1mm][c]{0.08\textwidth}{\centering\rotatebox[origin=b]{77}{#2}}}

\begin{document}
\maketitle
\begin{abstract}
    In a previous paper, 
    we proposed an introduction to the explainability of Graph Spectral Clustering results for textual documents, given that document similarity is computed as cosine similarity in term vector space. 
    In this paper, we generalize this idea by considering other embeddings of documents, in particular, based on the GloVe embedding 
    idea. 
\\\emph{Keywords:} 
Graph Spectral Clustering;
Explainable Clustering;
Explainable Artificial Intelligence;
Graph Spectral Clustering;
Word Embeddings (GloVe);
Textual Document Clustering
\end{abstract}

\section{Introduction}

The clustering of textual documents finds multiple practical applications. It is applied to detect relations between different texts, to distribute them dynamically in natural groups, or to discover the most relevant subjects within their content, and to express them in their own terms.
It is used for automatic document organization, topic extraction, and fast information retrieval and filtering. 

An approach to text clustering involves embedding documents in a space with well-defined properties and applying common algorithms, such as $k$-means \cite{Lloyd:1982}, for clustering in this space. 

The first embedding space used is the term vector space \cite{Salton:1975}, \cite{Bisht:2013}.
Clustering in term vector space has a big advantage, as it is quite simple to explain cluster membership by describing cluster centers with the highest weighing terms. 
The problem with such an embedding is that the dimensionality can be as high as dozens of thousands, even for a moderate-sized document collection of several thousand. Fortunately, if Graph Spectral Clustering (GSC) is used when clustering into several clusters, an embedding space with only a few dimensions might be adequate.

However, GSC based clustering result is hard to explain as the relationship between the embedding space coordinates and document words/terms gets lost. 
Happily, recently, a methodology was found to explain cluster membership in GSC in terms of document space words/terms \cite{Plosone2025}.

Although the term vector space is conceptually a simple construct, it has a significant disadvantage to text documents because the document is treated as a collection (or ``bag'') of words, so the context and relationships between terms are lost.
Therefore, new embedding approaches, such as Word2Vec \cite{Rong:2014:word2vec}, Doc2Vec \cite{lau2016empiricaldoc2vec}, GloVe \cite{Pennington:2014:glove}, or BERT \cite{devlin2019bert}, and many others, were developed to take such relationships into account. 
As a result, each word of a document and the document itself, is embedded in a 100 to 1,000-dimensional space in which the cosine similarity reflects the semantic similarity. 

Although the space dimensionality in such embeddings is greatly reduced, so that e.g., $k$-means clustering in such a space would be more efficient, applying GSC can greatly simplify the clustering process.  
Still, there exists a cluster explanation (clarification) problem, as not only do the cluster center coordinates in the GSC result have nothing to do with the document terms, but even the relationship between the cluster center in, say, a GloVe embedding and the document terms remains unclear.

In this paper, we intend to overcome this difficulty by proposing a method of explanation of clustering for GSC based on similarities computed as cosine similarities between GloVe embeddings by fusion of information from the GloVe embedding, original documents, and GSC analysis. 

In particular, in section \ref{sec:GloVeEmbedClust} 
we describe methodology of clustering under GloVe Vector Embedding. 
In section \ref{sec:GloVeExplanation} we show that   GloVe Vector Space Clustering explanation friendly.
In section \ref{sec:wGloVeEmbedClust} we extend the clustering considerations to a particular way of document weighting. 
In section \ref{sec:wGloVeExplanation} we add the explanation component.  
In section \ref{sec:GSC_GloVe} we explain how Graph Spectral Clustering (GSC) may wotk under   GloVe document embedding. 
In sections \ref{sec:Kclustering}, \ref{sec:LeqK},  \ref {sec:KeqGloVe} we show that clustering in GSC  based on combinatorial Laplacian  and clustering in GloVe document embedding deliver approximately same results so that the results from GSC based on combinatorial Laplacian  can be explained in a manner identical to explanation of results of clustering in  GloVe document embedding.  
Sections \ref{sec:Bclustering}, \ref{sec:NRLeqB}, \ref{sec:NRLeqB}, \ref{sec:BeqwGloVe} show that clustering in GSC  based on normalized Laplacian  and clustering in weighted GloVe document embedding deliver approximately same results so that the results from GSC based on normalized Laplacian  can be explained in a manner identical to explanation of results of clustering in  GloVe document embedding.  
In section \ref{sec:experiments} we present some experimental results on mentioned approximate clustering equivalence and show an example of clustering explanations. 
The paper ends with section \ref{sec:future} discussing possibilities of extension of the results to other new document embeddings.

\subsection{Highlights}
\begin{itemize}
\item In this research we perform a fusion of information immanent for Graph Spectral Clustering and for GloVe short document embedding leading to an enrichment of document similarity information present in term vector space with term interrelationship information. 
\item 
We show experimentally, that the term vector space embeddings are more advantageous than GloVe embeddings for short documents like those in Twitter data.   One may suspect that it is due to the sparseness of text in the
tweets and/or the dictionary limitations, excluding numerous words used in
the tweets. On the other hand the interrelationships of terms do not compensate the losses.  The worst results are achieved using GloVe embedding trained on Twitter data. A probable explanation may be that it contains a large amount of trash. 
\item 
We are extending the range of Explainable Graph Spectral Methods (Explainable GSC) from the domain of term vector space similarities to document similarities in GloVe embedding.
\item Explainability of Graph Spectral Clustering is achieved via building a bridge of (nearly) equivalent embeddings from GloVe embedding to GSC embedding.
\item We demonstrate on an example that an explanation built in the GloVe embedding may turn out to be more appealing than in term vector space embedding.
\end{itemize}

\section{GloVe Vector Embedding and Clustering}
\label{sec:GloVeEmbedClust}

GloVe embedding\footnote{
Elaborated at Stanford University, 
\url{https://nlp.stanford.edu/projects/glove/}
}  assigns the word $w$ a 100- or 300-dimensional vector of real numbers (or of other dimensionality), $g(w)$, so that similar words are close in this space.  
Although GloVe does not foresee a methodology for assigning a proper vector to a document, a linear transformation is likely to work (see, e.g. \cite{Nagoudi:2017}, Section 3.3.1). 
Several other authors follow this path (called BOW) \cite{Socher:2013}, \cite{Banea:2014:simcompass}, \cite{Hu:2014}.

This section deals with clustering via $k$-means of documents with GloVe embedding. 
We will present formulas expressing cluster centers in terms of words occurring in the cluster documents - eq. \eqref{eq:cc_infunct_of_words}. 
 
The most primitive assumption is to give equal weight to each word of the document. 

\begin{equation}
   g'(\dok)=\frac{1}{length(\dok)} \sum_{w \in \dok} g(w)   
\end{equation}
$g'(\dok)$ will be normalized to $g(\dok)$ so that $g(\dok)$ is of unit length. For any vector $v$ let $\alpha(v)=\frac{1}{\|v\|} $. Define 

\begin{equation}
   g(\dok)=\frac{\alpha(g'(\dok))}{length(\dok)}  \sum_{w \in \dok} g(w)   
\end{equation}
After this embedding, run the $k$-means algorithms to cluster the documents. 
Each cluster $C$ will have a cluster center {(or prototype)} $\mu(C)$ such that 
\begin{equation}\label{eq_center}
  \mu(C)= \frac{1}{|C|} \sum_{\dok \in C} g(\dok)   
\end{equation}

 A more sophisticated approach would be to weight the terms in the document, for example, with $tf$, $idf$, \textit{tf-idf}, or some other word weighting scheme. 
 Recall that $tf(w,\dok)$ means the number of occurrences of $w$ in $\dok$. 
 $df(w,\DokSet)$ is the number of documents $\dok\in \DokSet$ that contain the word $w$.
 Then $idf(w,\DokSet)= \frac{{|\DokSet|}}{\log(df(w,\DokSet))+1}$ if any $\dok\in \DokSet$ contains the word $w$, and is equal to 0 otherwise. 
 $tfidf(w,\dok,D)=tf(w,\dok) \cdot idf(w,\DokSet)$.  
 $tf, idf, tfidf$ can be generalized to a weighting function $weight(w,\dok,\DokSet)$. 
Then    
 
\begin{equation}
   g'(\dok)=\frac{1}{\sum_{w \in\dok } weight(w,\dok,\DokSet )} \sum_{w \in\dok } weight(w,\dok,\DokSet ) g(w)   
\end{equation}
$g'(\dok)$ will be normalized to $g(\dok)$ so that $g(\dok)$ is of unit length.

\begin{equation}
   g(\dok)= \alpha(g'(\dok))  \frac{ \sum_{w \in\dok } weight(w,\dok,\DokSet ) g(w)}   
   {\sum_{w \in\dok } weight(w,\dok,\DokSet )} 
\end{equation}

Let us denote

\begin{equation}
   weight^*(w,\dok,\DokSet ) =   \frac{ \alpha(g'(\dok))   weight(w,\dok,\DokSet )  }   
   {\sum_{w \in\dok } weight(w,\dok,\DokSet )} 
\end{equation}
Then 
\begin{equation}
   g(\dok)=     
   \sum_{w \in\dok } weight^*(w,\dok,\DokSet ) g(w) 
\end{equation}

So, the impact of a single word embedding on the document embedding amounts to:

\begin{equation}
   impact(w;\dok)=      
   \frac{\alpha(g'(\dok))}{\sum_{w \in\dok } weight(w,\dok,\DokSet )} weight(w,\dok,\DokSet )
   =   weight^*(w,\dok,\DokSet )
\end{equation}

Let us consider how similar the word is to the document that contains it. 
\begin{equation}
   sim(w,d)= impact(w;d) g(w)^T g(\dok)     
\end{equation}
Such a similarity measure reflects two things: 
on the one hand, the frequency of the word in the document, on the other hand, an indirect influence on the document via similarity to other words in the document. 

Notably  
\begin{equation}
\begin{split}
 \sum_{w\in\dok }  sim(w,d)
&=  \sum_{w \in\dok } impact(w;d) g(w)^T g(\dok)
\\&=  \sum_{w \in\dok }  g(\dok)^T impact(w;d) g(w)
\\&= g(\dok)^T  \left(\sum_{w\in\dok } impact(w;d) g(w) \right)
\\&=  g(\dok)^T  \left(\sum_{w\in\dok }  \frac{\alpha(g'(\dok))}{\sum_{w \in\dok } weight(w,\dok,\DokSet )} weight(w,\dok,\DokSet ) g(w)  \right)
\\&=  g(\dok)^T \alpha(g'(\dok)) \frac{\sum_{w\in\dok }   weight(w,\dok,\DokSet ) g(w) }{\sum_{w \in\dok } weight(w,\dok,\DokSet )}  
\\&=  g(\dok)^T \alpha(g'(\dok)) g'(\dok)    
\\&=  g(\dok)^T   g(\dok)=\|g(\dok)\|^2    
\\ &= 1      
\end{split}
\end{equation}

After this embedding, run $k$-means algorithms to cluster the documents. 
Each cluster $C$ will have a cluster center $\mu(C)$ 
 {as defined in (\ref{eq_center}). }  {It can be reformulated as}

\begin{equation}
  \mu(C)= \frac{1}{|C|} \sum_{\dok \in C} 
   \sum_{w \in\dok } weight^*(w,\dok,\DokSet ) g(w)
\end{equation}

\begin{equation}
  \mu(C)= \frac{1}{|C|} \sum_{w \in C} 
   \sum_{\dok; \dok\in C,  w \in\dok } weight^*(w,\dok,\DokSet ) g(w)
\end{equation}

\begin{equation}
  \mu(C)= \frac{1}{|C|} \sum_{w \in C} g(w)
   \sum_{\dok; \dok\in C,  w \in\dok } weight^*(w,\dok,\DokSet ) \label{eq:cc_infunct_of_words}
\end{equation}

So, the impact of a single word embedding on the cluster center embedding amounts to:

\begin{equation}
   impact(w;C)= 
    \frac{1}{|C|}  
   \sum_{\dok; \dok\in C,  w \in\dok } weight^*(w,\dok,\DokSet ) 
\end{equation}

\section{Why is GloVe Vector Space Clustering Explanation Friendly}
\label{sec:GloVeExplanation}

In this section, we show how the cluster center of a set of documents clustered via $k$-means under GloVe embedding of documents can be explained in terms of words from the documents.  That is by taking the most similar words, eq. \eqref{eq:wordCluster:sim}, or the most discriminating words, eq. \eqref{eq:wordCluster:diff:sim}.  

In the term vector space, finding the impact of a single word on the cluster center is simple: just take the coordinate of the cluster center related to the given word and it is done. It is as if words had coordinates in the term vector space such that all word coordinates are zeros except for a single one. And for each word the non-zero component is a  coordinate distinct from all the other words. So each word embedding vector is orthogonal to all the other. In the GloVe embedding, it is not that simple. Multiple words have same non-zero coordinates which reflects the fact that words may have related meanings. This introduces, however, the complication that word embedding vectors are not orthogonal. But on the other hand the coordinates of a document in the space are not only a function of the words occurring in the document, but also of the general word similarities. Hence the position of a document in the GloVe space is a fusion of the document content and the geneal knowledge carried by the GloVe system.   

We can compute, for each word, its similarity to $\mu(C)$ as follows:
\begin{equation} \label{eq:wordCluster:sim}
    sim(w,C)=impact(w;C)) g(w)^T \mu(C)
\end{equation}
Note that such an expression for similarity is justified also for term vector space. But in the term vector space, the expression would reduce to considering only one coordinate of $g(w)$ and $ \mu(C)$, while the contribution of the other would be zero. So only a concrete word occurrence in each document would play a role in TVS. 
It is different for GloVe. 

Modifications like $tf$, \textit{tf-idf} are of course encompassed in $weight^*(w,\dok,\DokSet )$. 
The above formula implies  that 
the contribution of $w$ to the cluster center depends not only on $w$ itself but also on similar words when they occur in the documents as the word embedding vectors are not orthogonal. 

One property, however, would be shared by both TVS and GloVe. 
Recall that if in TVS a similarity of a word to the cluster center is considered as the square of the respective cluster center coordinate, then the sum of these similarities is the squared length of the cluster center vector.  

It turns out that under the similarity definition in eq. \eqref{eq:wordCluster:sim}, the same property holds also for GloVe. See below: 
\begin{align}
 \sum_{w\in C}  sim(w,C)
 &=   \sum_{w\in C} impact(w;C) g(w)^T \mu(C)
\\ &=   \sum_{w\in C} impact(w;C) \mu(C)^T g(w) 
\\ &=  \mu(C)^T \sum_{w\in C} impact(w;C)  g(w) 
\\ &=  \mu(C)^T \sum_{w\in C} \frac{1}{|C|}  
   \sum_{\dok; \dok\in C,  w \in\dok } weight^*(w,\dok,\DokSet ) 
  g(w) 
\\ &=  \mu(C)^T  \frac{1}{|C|} \sum_{w\in C}   g(w)  
   \sum_{\dok; \dok\in C,  w \in\dok } weight^*(w,\dok,\DokSet ) 
 \\ &= \mu(C)^T \mu(C)   = \|  \mu(C) \|^2   
\end{align}

Let us consider the issue of explaining why a document belongs to a cluster. 
Take the squared distance between a document and a cluster center:
$\|g(\dok)-\mu(C)\|^2
=g(\dok)^Tg(\dok)
-2g(\dok)^T\mu(C)
+\mu(C)^T\mu(C)
$.
Here, $g(\dok)^Tg(\dok)$ and $\mu(C)^T\mu(C)$ may be treated as those that do not say anything about the membership of the cluster. 

So, the distance is driven by $-g(\dok)^T\mu(C)$, hence the similarity may be expressed as $g(\dok)^T\mu(C)$. 
\begin{align}
    g(\dok)^T\mu(C)=
    \sum_{w \in\dok } weight^*(w,\dok,\DokSet ) g(w)^T\mu(C)
\end{align}
So the contribution of a single word to the similarity of a document to the cluster center amounts to
\begin{align} 
         weight^*(w,\dok,\DokSet ) g(w)^T\mu(C) 
\end{align}

Sort the values of this quantity for all the words in the document, and select the top ones to identify which words most significantly contribute to cluster membership.

The aforementioned explanation of clusters via the terms may be enhanced by differentiating it from other clusters.     
Consider the cluster $C$ and the set $\mathcal{C}$ of all other clusters in the given clustering. One of the measures of distinction of a cluster can be the sum of squared distances to other clusters.    
\begin{equation}\label{eq:clusterversusotherclustersGloVeX}
   ClDiff(C)= \sum_{C'\in \mathcal{C}}   
   \|\mu(C)-\mu(C')\|^2
\end{equation}
which is the sum of squared differences between the given cluster and the other clusters.  It can be rewritten as
\begin{align}\label{eq:clusterversusotherclustersGloVe}
   ClDiff(C)= & \sum_{C'\in \mathcal{C}} \mu(C)^T\mu(C)  
   + \sum_{C'\in \mathcal{C}} \mu(C')^T\mu(C')
   - 2\sum_{C'\in \mathcal{C}} \mu(C)^T\mu(C')\\
   = &  
    \sum_{C'\in \mathcal{C}} \|\mu(C')\|^2 
   + |\mathcal{C}|   \|\mu(C)\|^2  
   - 2 \mu(C)^T\sum_{C'\in \mathcal{C}}\mu(C') \label{eq:ClDiff}
\end{align}
We can now think of cluster distinction as an attempt to move our cluster away from the other.
This moving away is deemed to be accomplished by adding more weight to some words. The words for which adding weight moves the cluster farther away can be considered the most important ones. In other words, the highest value on the first derivative with respect to the word is an indicator of the highest importance of the word for distinguishing this cluster from the others. Here, we assume that we do not change the weights of the words in other clusters, we consider the manipulation of one cluster at a time. 
Thus, for each $C'\in \mathcal{C}$, the vector  $\mu(C')$ remains unchanged.  See %
{\eqref{eq:ClDiff}}.
 So driving the distance towards maximum would mean maximizing  
   $ \mu(C)^T( |\mathcal{C}| \mu(C)-2\sum_{C'\in \mathcal{C}}\mu(C'))$.
The impact of changing the presence of a single word $w$ would then amount to
\begin{align}
    \frac{d}{dw} & \left( \mu(C)^T\left( |\mathcal{C}| \mu(C)-2\sum_{C'\in \mathcal{C}}\mu(C')\right)\right)
    \\ = &
    \left(\frac{d\mu(C)^T}{dw}\right) \left( |\mathcal{C}| \mu(C)-2\sum_{C'\in \mathcal{C}}\mu(C')\right) 
   \\ &+
      \mu(C)^T\frac{d}{dw}\left( |\mathcal{C}| \mu(C)-2\sum_{C'\in \mathcal{C}}\mu(C')\right)
    \\ = &
    \left(\frac{d\mu(C)^T}{dw}\right) \left( |\mathcal{C}| \mu(C)-2\sum_{C'\in \mathcal{C}}\mu(C')\right) 
    +
     |\mathcal{C}| \mu(C)^T\frac{d  \mu(C)}{dw} 
    \\ = &
    \left(\frac{d\mu(C)^T}{dw}\right) \left( |\mathcal{C}| 2\mu(C)-2\sum_{C'\in \mathcal{C}}\mu(C')\right) 
\end{align}

Note that 
   \begin{align}
   \frac{d  \mu(C)}{dw}& =
   \frac{d  }{dw} \left(
 \frac{1}{|C|} \sum_{w \in C} 
   \sum_{\dok; \dok\in C,  w \in\dok } weight^*(w,\dok,\DokSet ) g(w)
   \right)
   \\ &=  \frac{1}{|C|}
  \sum_{\dok; \dok\in C,  w \in\dok } weight^*(w,\dok,\DokSet )  g(w)  \end{align}
Hence each word $w$ contributes strength \\
{\footnotesize  
\begin{align}
    \left(\frac{1}{|C|}
  \sum_{\dok; \dok\in C,  w \in\dok } weight^*(w,\dok,\DokSet )   \right) g^T(w) \left( |\mathcal{C}| 2\mu(C)-2\sum_{C'\in \mathcal{C}}\mu(C')\right) 
  \\=
    2\frac{(|\mathcal{C}|+1)}{|C|}
  \left(\sum_{\dok; \dok\in C,  w \in\dok } weight^*(w,\dok,\DokSet )   \right) g^T(w) \left( \mu(C)-
  \frac{1}{(|\mathcal{C}|+1)}
  \sum_{C'\in \mathcal{C}\cup \{C\}}\mu(C')\right) 
  \\=
  \label{eq:wordCluster:diff:sim}
    2 (|\mathcal{C}|+1) 
  impact(w;C)  g^T(w) \left( \mu(C)-
  \frac{1}{(|\mathcal{C}|+1)}
  \sum_{C'\in \mathcal{C}\cup \{C\}}\mu(C')\right) 
\end{align}
}

If we sort by this quantity, the words with the top impact would explain the cluster in contrast to the other clusters.

\section{Weighted GloVe Vector Embedding and Clustering}
\label{sec:wGloVeEmbedClust}

This section introduces weighted clustering of documents in the GloVe embedding. 
The weighting has two purposes: to draw towards coordinate GloVe system origin those documents which are most similar to other documents (can be seen as kinds of representatives) and to put more weight on such documents while clustering. 

Consider the $g(\dok_i)$  {unit} vectors representing embedding in GloVe space  introduced in Section \ref{sec:GloVeEmbedClust}.
In the sequel we will use vectors of the form 
\begin{equation}
g_\omega(\dok_i)=g(\dok_i)/\omega_i    
\end{equation}
where $\omega_i$ is defined as
$$\omega_i=\sum_\ell s_{i\ell}$$
or \footnote{For details see Section \ref{sec:GSC_GloVe}.}
$$\omega_i=1+\sum_\ell s_{i\ell}$$

Denote by $\Gamma$ a partition of the set of documents into $k$ disjoint groups, $\Gamma=\{C_1,\dots,C_k\}$. One can perform weighted $k$-means clustering    {as defined  e.g. in \cite{Dhillon:2004})}:
\begin{align}  
Q^{[{\omega}GloVe]}(\Gamma; \boldsymbol\omega)
& =\sum_{C\in \Gamma} \sum_{\dok_i \in C} \omega_i  
\|g_\omega(\dok_i)-\mu(C)\|^2   
\\& 
=\sum_{j=1}^k \frac{1}{V_j} \sum_{\dok_i, \dok_\ell \in C_j} \omega_i \omega_\ell
\|g_\omega(\dok_i)-g_\omega(\dok_\ell)\|^2   \label{eq:NRwGloVeminimization}
\end{align}
where $V_j=\sum_{\ell \in C_j} \omega_\ell$ stands for the ``volume'' of $j$-th cluster.

For any cluster $C_j \in \Gamma$
\begin{equation}
  \mu(C_j)= \frac{1}{\Vol_j} \sum_{\dok_i \in C_j} 
  \omega_i g_\omega(\dok_i)
  = \frac{1}{\Vol_j} \sum_{\dok_i \in C_j} 
   g(\dok_i)
\end{equation}

 {This formula resembles (\ref{eq_center}), except that the cardinality $|C_j|$ has been replaced by the volume $V_j$. This results in shifted centers, and in different cluster membership.} Anyway, the impact of individual words is accounted for in the same way as for unweighted cases.

So, the impact of a single word embedding on the cluster center embedding amounts to:

\begin{equation}
   impact(w;C_j)= 
    \frac{1}{\Vol_j}  
   \sum_{\dok; \dok\in C,  w \in\dok. } weight^*(w,\dok,\DokSet ) 
\end{equation}

\section{Why is weighted GloVe Vector Space Clustering Explanation Friendly}
\label{sec:wGloVeExplanation}
All the considerations affecting the explanation for unweighted GloVe vector space explanations apply here accordingly. 

In particular, the explanation of the cluster center is formally the same. 
Also that of document membership. The difference occurs when handling a cluster versus other clusters. Then not cardinalities but volumes are considered.

\section{GSC with combinatorial and other Laplacian Clustering and GloVe document embedding}
\label{sec:GSC_GloVe}

This section explains how to exploit the properties of Graph Spectral Clustering methodology when clustering documents in GloVe embedding. 
Using of GSC for clustering instead of direct clustering in the GloVe embedding has the advantage of reducing the dimensionality of the clustering problem by an order of magnitude. 
As we will show in subsequent sections, 
the result of GSC approximates the result of direct clustering in GloVe embedding. In this way, it has the property of lower dimensionality for clustering as GSC and the property of explainability as GloVe embedding.

Graph spectral clustering methods  are a relaxation of cut-based graph clustering methods.  {Below we briefly characterize this approach. A more detailed description can be found, for example, in \cite{vonLuxburg:2007}.}

Let $S$ be a (symmetric) similarity matrix between pairs of items (documents, in our case). It induces a graph whose nodes correspond to the entities (documents). Let $n$ denote the number of items for which $S$ has been computed. Let $D$ be the diagonal matrix with $d_{jj}=\sum_{k=1}^n s_{jk}$ for each  
$j \in [n]$
In the domain of text mining, the similarity matrix is usually based on either a graph representation of relationships (links) between items  or such a graph is induced by (cosine) similarity measures between these items.   
However, mixed object representations (text and links) have also been studied \cite{Xu:2016}. 
 {Here we study a special kind of similarity measure, namely the cosine similarity between GloVe embedding vectors.}

Consider two documents $d_i, d_\ell$ and their embeddings $g(\dok_i), g(\dok_\ell)$, which by definition are of unit length. The $(i,j)$-th entry of the similarity matrix is equal to $s_{i\ell}= g(\dok_i)^T g(\dok_\ell)$. By convention, all diagonal elements of the matrix $S$ are zero. 
A(n unnormalised or) combinatorial Laplacian $L$ corresponding to this matrix (approximating the RCut) is defined as 
\begin{equation}\label{eq:combLapDef} L=D-S, \end{equation}
where $D$ is the diagonal matrix with $d_{ii}=\sum_{\ell=1}^ns_{i\ell}$ for each $i \in [n]$. 
A \emph{normalized Laplacian} $\mathcal{L}$ of the graph represented by~$S$ is defined as 
\begin{equation}\label{eq:normLapDef}\mathcal{L}=D^{-1/2}L D^{-1/2}= I -D^{-1/2}S D^{-1/2} .\end{equation}%
The \emph{rationormalized Laplacian}%
\footnote{
Other Laplacians are also used, e.g. the random walk Laplacian $\mathbb{L}$  of a graph, defined as 
\begin{equation}\mathbb{L}=LD^{-1}  = I -SD^{-1} \end{equation}
Other Laplacians were also studied~\cite{STWMAKSpringer:2018}. 
}
 {takes the form} 
\begin{align}\label{eq:ratnormLapDef}\mathcal{L_R}=&{D'}^{-1/2}L {D'}^{-1/2}= 
I -{D'}^{-1/2}S' {D'}^{-1/2} 
\end{align}%
where $S'=S+I$ and $D'=D+I$.

The division into $k$ disjoint clusters is done as follows. 
One computes the eigendecomposition of the respective Laplacian,
getting $n$ eigenvalues $\lambda_1\le\dots\le\lambda_n$ (always $\lambda_1=0$) and corresponding eigenvectors $\mathbf{v}_1,\dots,\mathbf{v}_n$. Then one embeds the documents into the $k$-dimensional space spanned by the $k$ eigenvectors corresponding to $k$ lowest eigenvalues. That is, $i$-th document is represented by the vector $\mathfrak{x}_i = [v_{i,2},\dots,v_{i,k+1}]^T$. This shall be called $L$-embedding if the eigenvectors are determined from the combinatorial Laplacian $L$, $N$-embedding if the normalized Laplacian is used, and $R$-embedding if the rationormalized Laplacian is used. Please note that the eigenvector $\mathbf{v}_1$ of the combinatorial Laplacian is a constant vector, hence to get informative embedding, we take into account the eigenvectors $\mathbf{v}_2,\dots,\mathbf{v}_{k+1}$. We also apply this convention to the other matrices $\mathcal{L}$ and $R$ for consistency. Note also that the $L$ embedding allows us to approximate the so-called RCut criterion. In contrast, the $N$-embedding approximates the NCut criterion, while $R$-embedding allows the approximation of the NRCut clustering criterion from the graph domain as described in  \cite{RozdzialvMonografia:2024}. See \cite{vonLuxburg:2007} or \cite{STWMAKSpringer:2018}) for details. 
Then one clusters the documents in a chosen embedding space using e.g. $k$-means algorithm. 

In summary, we can formulate the three criteria of the form 
\begin{equation}\label{eq:Xminimization} 
Q^{[GSA \mathfrak{X}]}(\Gamma)=\sum_{j=1}^k  \sum_{i \in C_j} ||\mathfrak{x}_i -\boldsymbol\mu(C_j)||^2  = \frac{1}{2}\sum_{j=1}^k \frac{1}{|C_j|} \sum_{i \in C_j}\sum_{\ell \in C_j}\|\mathfrak{x}_i - \mathfrak{x}_\ell\|^2 
\end{equation}
where $\Gamma=\{C_1,\dots, C_k\}$ is a partition of the set of documents, $\mathfrak{X} \in \{L,\mathcal{L}, R\}$, and $\mathfrak{x}_i$ stands for a vector consisting of the $i$-th entries of the $k$ eigenvectors of the matrix $L$, if $\mathfrak{X}=L$, or the $i$-th positions of the $k$ eigenvectors of the matrix $\mathcal{L}$, if $\mathfrak{X}=\mathcal{L}$, or the eigenvectors of the matrix $R$, if $\mathfrak{X}=R$.  

The $j$-th cluster center is defined  as in equation \eqref{eq_center}, i.e.
\begin{equation}\label{eq:Q::kmeans:original:center}
 \boldsymbol\mu(C_j)=\frac1{|C_j|} \sum_{i \in C_j} \mathfrak{x}_i 
\end{equation}
where $\mathfrak{x}_i$ represents embedding of $i$-th document as described above.  

The formula \eqref{eq:Xminimization}, is exactly the target function of the $k$-means algorithm in the respective (Euclidean) embedding space. It is hence quite natural that their minima are sought using the traditional $k$-means algorithm.   

In the context of graphs clustering, the main quantity is 
\begin{equation}
    cut(C_j,\bar{C}_j) = \sum_{i \in C_j}\sum_{\ell \notin C_j} s_{i\ell}
\end{equation}
It represents the aggregate similarity of nodes that are neighbors in a given graph, but belong to different clusters. Properly normalizing this quantity we obtain three different criteria, i.e.
\begin{subequations}\label{cuts}
\begin{align}
\label{eq:qRCut_def}
RCut(\Gamma) = \sum_{j=1}^k \frac{cut(C_j,\bar{C}_j)}{|C_j|}=\sum_{j=1}^k \frac{1}{|C_j|}\sum_{i \in C_j}\sum_{\ell \notin C_j}s_{i\ell}\\
NCut(\Gamma) = \sum_{j=1}^k \frac{cut(C_j,\bar{C}_j)}{\mathcal{V}_j}=\sum_{j=1}^k \frac{1}{\mathcal{V}_j}\sum_{i \in C_j}\sum_{\ell \notin C_j}s_{i\ell}\\
\label{eq:qNRCut_def}
NRCut(\Gamma)= \sum_{j=1}^k \frac{cut(C_j,\bar{C}_j)}{\mathcal{V}'_j}=\sum_{j=1}^k \frac{1}{\mathcal{V}'_j}\sum_{i \in C_j}\sum_{\ell \notin C_j}s_{i\ell}
\end{align}
\end{subequations}
where 
\begin{equation}
\Vol'_j = \sum_{i \in C_j} (d_{ii}+1) = \Vol_j + |C_j| 
\end{equation} 
Thus the NRCut clustering can be viewed as a mixture of NCut and RCut. 

Note the symmetry between the average $k$ criterion and the cutting criterion. In \eqref{eq:Xminimization}, the average difference (measured by the Euclidean distance) between the members of the $j$-th cluster is minimized, while in \eqref{cuts}, the average similarity between neighboring nodes assigned to different clusters is minimized, respectively. 
\section{\texorpdfstring{$K$}{K}-embedding   (For Use With \texorpdfstring{$k$}{k}-Means)}\label{sec:Kclustering}

Recall a new embedding of the documents from $\mathcal{D}$,
 introduced in \cite{Plosone2025}. 
Let $A$ be a matrix of the form:
\begin{equation}
    A= \mathbf{1}\mathbf{1}^T-I-S
\end{equation}
\noindent where $I$ is the identity matrix, and $\mathbf{1}$ is the (column) vector consisting of ones, both of appropriate dimensions. 
The matrix $A$ is nonnegative and has a diagonal equal to zero, so it may be considered as a kind of (squared) pseudo-distance, needed by the Gower embedding method \cite{Gower:1966}, used below.  As mentioned in the previous section, the diagonal of $S$ consists of zeros.
Let $K$, hereafter referred to as the $K$ matrix, be a  {doubly centered matrix of order~$n$ defined as follows}, \cite{Gower:1966} 
\begin{equation}
    K=-\frac12(I-\frac1{n}\mathbf{1}\mathbf{1}^T)A(I-\frac1{n}\mathbf{1}\mathbf{1}^T)\;,
\end{equation}
\noindent  Since $(I-\frac1{n}\mathbf{1}\mathbf{1}^T)\mathbf{1}
=\mathbf{1}-\frac1{n}\mathbf{1}\mathbf{1}^T\mathbf{1}
=\mathbf{1}-\frac1{n}\mathbf{1}n=\mathbf{0}$, we state that 
 is an eigenvector of $K$.  Since $K$ is a real and symmetric matrix, all its other eigenvectors must be orthogonal to $\mathbf{1}$, i.e. $\mathbf{1}^T\mathbf{v}=0$.

Let $\Lambda$ be a diagonal matrix of $K$ eigenvalues, and $V$ a matrix whose columns are the corresponding $K$ eigenvectors (of unit length).
Then $K=V\Lambda V^T$. 
Let $\mathbf{z}_i=\Lambda^{1/2} V^T_{i}$, where $V_i$ stands for $i$-th row of $V$.  
Let $\mathbf{z}_i,\mathbf{z}_\ell$ be the {embedding}s of the documents $i,\ell$, resp. 
This embedding shall be called \emph{$K$-embedding}. 
Then (see \cite{RAKMAKSTW:2020:trick}) 
\begin{equation}    \label{eq:embeddist}
\|\mathbf{z}_i-\mathbf{z}_\ell\|^2=
 A_{i\ell}= 1-s_{i\ell}
\end{equation}
for $i\ne \ell$. Hence, upon performing $k$-means clustering in this space, we \emph{de facto} try to maximize the sum of similarities within a cluster. 
\footnote{
Lingoes correction is needed, if $K$ turns out to have negative eigenvalues, see \cite{RAKMAKSTW:2020:trick}. 
The correction consists in adding $2\sigma$ to all elements of dissimilarity matrix $A$ except for the main diagonal, which has to stay equal to zero, where $\sigma\ge -\lambda_m$  where $\lambda_m$ is the smallest eigenvalue of $K$. By adding, we get a new matrix $A'$, for which we compute new $K'$ and use the prescribed embedding resulting from $K'$ and not from $K$, when performing $k$-means. }

The above embedding can be used for clustering documents using the $k$-means algorithm. The minimized quality function has a form
\begin{equation}\label{eq:Kminimization}
Q^{[Kbased]}(\Gamma)=\sum_{j=1}^k \sum_{i\in C_j} ||\mathbf{z}_i-\boldsymbol\mu(C_j)||^2  
=\sum_{j=1}^k \frac{1}{2|C_j|} \sum_{i \in C_j} 
\sum_{\ell \in C_j} \|\mathbf{z}_i - \mathbf{z}_\ell\|^2\;  
\end{equation}
where the prototype $\boldsymbol\mu(C_j)$ is defined as in \eqref{eq:Q::kmeans:original:center}.
This implies  
\begin{align*}
Q^{[Kbased]}(\Gamma)=& %
\sum_{j=1}^k \frac{1}{2|C_j|} \sum_{i \in C_j} 
\sum_{\substack{\ell \in C_j\\ \ell \ne i}} 
(1-s_{i\ell})%
\end{align*}%
that is  
\begin{equation} \label{eq:target:Kbased} 
Q^{[Kbased]}(\Gamma)=
\frac{n-k}2-\sum_{j=1}^k \frac{1}{2|C_j|} \sum_{i \in C_j} 
\sum_{\ell \in C_j; \ell\ne i} 
s_{i\ell}\;
\end{equation}
where 
$n,k$ are independent of clustering. 

Instead of using all eigenvectors to represent the $K$, the top $m$ eigenvalues and associated eigenvalues can be used to approximate it sufficiently.

\section{Relationship Between \texorpdfstring{$L$}{L}-based Clustering and \texorpdfstring{$K$}{K}-based Clustering}\label{sec:LeqK}
We see from eq. \eqref{eq:target:Kbased} that $k$-means applied to $K$-based embedding seeks to find the clustering that maximizes the sum of the average similarity within a cluster. 
On the other hand, the intention of the $L$-based GSA described previously in  {Sec.}\ref{sec:GSC_GloVe} is to approximate the target of RCut, which is expressed via formula \eqref{eq:qRCut_def},  is to minimize the sum of average similarity to elements of other clusters. 
These goals are similar, but not identical, as we shall see. 
Let us compute the difference between them: 
\begin{align*}
      Q^{[RCut]}(\Gamma) - & 2Q^{[Kbased]}(\Gamma) \\ 
= &     
  \sum_{j=1}^k \frac{1}{|C_j|}  \sum_{i\in C_j} \sum_{\ell\not\in C_j}  s_{i\ell}
  -(n-k)
  +\sum_{j=1}^k \frac{1}{|C_j|} \sum_{i \in C_j} 
  \sum_{\ell \in C_j; \ell\ne i} s_{i\ell}\\
= & 
-(n-k) +\sum_{j=1}^k \frac{\Vol_j}{|C_j|}  
\end{align*}
 {If all target clusters had the same size, the above expression would be a constant, so optimizing any quality function would produce the same result. }
Otherwise, the same results can be achieved under the following assumption: 
The similarities within the good clusters are above some level $g$, and those between elements of different clusters are below some level $b$. If $g/\max_j(|C_j|)>b/\min_j(|C_j|)$, then also the optimal clustering of both is the same (if the smallest cluster is large).   
If these criteria are approximately matched, the optimal values will also be approximately the same.

\section{Equivalence Between \texorpdfstring{$K$}{K}-based Clustering and GloVe Embedding Space  Vector Based Clustering}\label{sec:KeqGloVe}

The following criterion can be minimised:
\begin{equation}\label{eq:GloVeminimization}
  Q^{[GloVe]}(\Gamma)=\sum_{j=1}^k \sum_{i\in C_j} ||g(\dok_i)-\boldsymbol\mu(C_j)||^2  
\end{equation}
 (with    {$\boldsymbol\mu(C_j)=\frac1{|C_j|} \sum_{i \in C_j} g(\dok_i)$ } )
 which may be reformulated as 
 \begin{equation} 
Q^{[GloVe]}(\Gamma)
=\sum_{j=1}^k \frac{1}{2|C_j|} \sum_{i \in C_j} 
\sum_{\ell \in C_j} \|g(\dok_i) - g(\dok_\ell)\|^2\;  
\end{equation} 
Obviously, for $i\ne \ell$
$$ \|g(\dok_i) - g(\dok_\ell)\|^2
=\|g(\dok_i)\|^2 
-2  g(\dok_i)^Tg(\dok_\ell)
+  \|g(\dok_\ell)\|^2
=2-2  s_{i\ell}
$$
This means that 
\begin{align}
Q^{[GloVe]}(\Gamma)
= &
\sum_{j=1}^k \frac{1}{|C_j|} \sum_{i \in C_j} 
\sum_{\ell \in C_j; \ell \ne i} (1-s_{i\ell})
\\ 
= &
n-k  
- 
\sum_{j=1}^k \frac{1}{|C_j|} \sum_{i \in C_j} 
\sum_{\ell \in C_j; \ell \ne i}
s_{i\ell}
\label{eq:target:GloVe}
\end{align}

A quick look at formulas \eqref{eq:target:Kbased} and
\eqref{eq:target:GloVe}
reveals that both clustering criteria are identical. 
Therefore clustering in $K$-based embedding and clustering in GloVe embedding optimize the same target function.
What is the difference? 
The $K$-embedding is lower dimensional, as the length of eigenvectors equals the number of documents. 
Here, the number of dimensions is equal to the richness of the embedding (100-300 dim), which may be 10 times as high or more. Hence the clustering under $K$-embedding will be significantly faster. 

What do we gain then via GloVe embedding?
We have already seen that under balanced-clusters 
clustering in $K$-based embedding approximates RCut clustering which on the other hand is approximated by $L$-based clustering. 
Therefore, the application of clustering and cluster membership explanation methods outlined in  {Sec.}\ref{sec:GloVeExplanation}, \ref{sec:GloVeEmbedClust} to the results of  $L$-based spectral clustering method is justified.

In summary, we have pointed out in this section that the traditional $L$-embedding lost the direct relation between {data}point distances and the cosine similarity of documents. This is a serious disadvantage because $k$-means is applied in GSA clusters based on distances in embedding space, not similarities between documents. 
We have shown that there exists a $K$ embedding having approximately the same general goal as $L$-embedding (see Sec.\ref{sec:LeqK}), but with the property that distances in the space are directly translated to similarities so that $k$-means applied in this embedding optimizes on the similarities within a cluster directly. 
In the third embedding, the GloVe embedding, the similarities can be computed directly as cosine similarity or based on Euclidean distances. This duality allows for precise pointing at sources of similarities of the cluster elements and at sources of dissimilarities in terms of words of the documents.  

In this way the problem of GSA explanation is overcome in that membership reason can be given in terms of sets of decisive words.

\section{\texorpdfstring{$B$}{B}-embedding   (For Use With \texorpdfstring{$k$}{k}-Means)}\label{sec:Bclustering}

Let us use the following notation:
$s'_{i\ell}=s_{i\ell}$ for $i\ne\ell$ and $s'_{ii}=1$,
$d'_{ii}=\sum_\ell s'_{i\ell}$,
${D'}$ be the matrix with diagonal $d'_{ii}$ and zeros elsewhere. 
$\omega'_{i}= d'_{ii}$, 
${\Vol}'_j=\sum_{i \in C_j} d'_{ii}$, 
$F_j=\sum_{i \in C_j} {d'}_{ii}^{-1}$, 
$F=\sum_{i \in \mathcal{D}} {d'}_{ii}^{-1}$, 
$\omega'_S=\sum_{i \in S} d'_{ii}$, 
$\omega'_{il}=\omega'_i \omega'_l$,

We suggest to use the $\mathcal{A}$ matrix of the following form, see \cite{RozdzialvMonografia:2024}. Denote by $\mathcal{E}$ a matrix of the form   
\begin{equation}
    \mathcal{E}= \mathbf{1}\mathbf{1}^T-I\;.
\end{equation}
Then define  
\begin{equation}
    \mathcal{A}= \mathcal{E}{D'}^{-2} + {D'}^{-2}\mathcal{E} -  2{D'}^{-1}S{D'}^{-1}\;.
\end{equation}
with $D',S$ being defined as previously.  
Let $\mathcal{B}$ be the matrix of the form:
\begin{equation}
    \mathcal{B}=-\frac12(I-\frac1{n}\mathbf{1}\mathbf{1}^T)\mathcal{A}(I-\frac1{n}\mathbf{1}\mathbf{1}^T)\;.
\end{equation}
We proceed with $\mathcal{B}$ in a similar way as with $K$ matrix. 
Note that $\mathbf{1}$ is an eigenvector of $\mathcal{B}$, with the corresponding eigenvalue equal to 0. 
All the other eigenvectors must be orthogonal to it as $\mathcal{B}$ is real and symmetric, so for any other eigenvector $\mathbf{v}$ of $\mathcal{B}$ we have: $\mathbf{1}^T\mathbf{v}=0$.

Let $\Lambda$ be the diagonal matrix of eigenvalues of $\mathcal{B}$, and $V$ the matrix where columns are corresponding (unit length) eigenvectors of $K$. 
Then $\mathcal{B}=V\Lambda V^T$. 
Let $\boldsymbol{\zeta}_i=\Lambda^{1/2} V^T_{i}$, where $V_i$ stands for $i$-th row of $V$.  
Let $\boldsymbol{\zeta}_i,\boldsymbol{\zeta}_\ell$ be the {embedding}s of the documents $i,\ell$, resp. 
This embedding shall be called \emph{$\mathcal{B}$-embedding}.
Then 
\begin{equation}    \label{eq:NRembeddistfrak}
\|\boldsymbol{\zeta}_i-\boldsymbol{\zeta}_\ell\|^2= \mathcal{A}_{i\ell}= 
\frac1{{d'}_{ii}^2}+\frac1{{d'}_{\ell\ell}^2}-2\frac{s_{i\ell}}{{d'}_{ii}{d'}_{\ell\ell}}
\end{equation}
Let us now discuss performing weighted $k$-means clustering on the vectors $\boldsymbol{\zeta}_i$ with weights amounting to ${d'}_{ii}$ respectively.

 \begin{equation} \label{eq:RNKminimization}
 Q^{[\mathcal{B} based]}(\Gamma; \boldsymbol{\omega'})=\sum_{j=1}^k \sum_{i\in C_j} \omega'_i\|\boldsymbol{\zeta}_i-\boldsymbol\mu_{\boldsymbol{\omega'}}(C_j)\|^2    
 \end{equation}
 whereby 
$$
 \boldsymbol\mu_{\boldsymbol{\omega'}}(C_j)
 =
 \frac{
 \sum_{\in C_j} \omega'_i\boldsymbol{\zeta}_i
 }
 {
 \sum_{\in C_j} {\omega'}_i
 }
 =\frac{1}{{\Vol'}_j}
 \sum_{\in C_j} \omega'_i\boldsymbol{\zeta}_i
 $$
 which may be reformulated as 

\Bem{
\begin{align} \label{eq:Q::NRkmeansW}
Q^{[\mathcal{B} based]}(\Gamma; \boldsymbol\omega')
&=\sum_{i=1}^n\sum_{j=1}^k u_{ij}\omega'_i\|\boldsymbol{\zeta}_i - \boldsymbol{\mu_\omega'}_j\|^2 \\
&=\sum_{j=1}^k \frac{1}{{\Vol'}_j} \sum_{\boldsymbol{\zeta}_i, \boldsymbol{\zeta}_\ell \in C_j} \omega'_i \omega'_\ell \|\boldsymbol{\zeta}_i - \boldsymbol{\zeta}_\ell\|^2 \\
&=\sum_{j=1}^k \frac{1}{2{\Vol'}_j} \sum_{i \in C_j} 
\sum_{\ell \in C_j} \omega'_i\omega'_\ell\|\boldsymbol{\zeta}_i - \boldsymbol{\zeta}_\ell\|^2\;  
\;  
\\&=\sum_{j=1}^k \frac{1}{2{\Vol'}_j} \sum_{i \in C_j} 
\sum_{\substack{\ell \in C_j\\ \ell \ne i}}
(\frac{{d'}_{ii}}{{d'}_{\ell\ell}}+\frac{{d'}_{\ell\ell}}{{d'}_{ii}}-2s_{i\ell}) 
\;  
\;  
\\&=\sum_{j=1}^k \frac{1}{2{\Vol'}_j} \sum_{i \in C_j} 
\sum_{\substack{\ell \in C_j}}
\left(\frac{{d'}_{ii}}{{d'}_{\ell\ell}}+\frac{{d'}_{\ell\ell}}{{d'}_{ii}}-2s'_{i\ell}\right) \label{eq:NRcmpwithGlove}
\;  
\;  
\\&=\sum_{j=1}^k \frac{1}{2{\Vol'}_j} \left(
\left(\sum_{i \in C_j}  \sum_{\substack{\ell \in C_j}} \frac{{d'}_{ii}}{{d'}_{\ell\ell}} \right)
+\left(\sum_{i \in C_j}  \sum_{\substack{\ell \in C_j}} \frac{{d'}_{\ell\ell}}{{d'}_{ii}} \right)
 \right. \\ & \left. 
-\left(\sum_{i \in C_j}  \sum_{\substack{\ell \in C_j}} 2s'_{i\ell} \right)
\right) \label{eq:NRQ40:tripledouublesums}
\;  
\\&=\sum_{j=1}^k \frac{1}{2{\Vol'}_j} \left(
\left(\sum_{i \in C_j} {d'}_{ii} \sum_{\substack{\ell \in C_j}} \frac{1}{{d'}_{\ell\ell}} \right)
 \right. \\ & \left. 
+\left(\sum_{i \in C_j} \frac{1}{{d'}_{ii}} \sum_{\substack{\ell \in C_j}} {d'}_{\ell\ell} \right)
-\left(\sum_{i \in C_j}  \sum_{\substack{\ell \in C_j}} 2s'_{i\ell} \right)
\right)  
\;  
\\&=\sum_{j=1}^k \frac{1}{2{\Vol'}_j} \left(
\left(\sum_{i \in C_j} {d'}_{ii} F_j \right)
+\left(\sum_{i \in C_j} \frac{1}{{d'}_{ii}} {\Vol'}_j \right)
 \right. \\ & \left. 
-\left(\sum_{i \in C_j}  \sum_{\substack{\ell \in C_j}} 2s'_{i\ell} \right)
\right)  
\;  
\end{align}
}
\begin{align}  
Q^{[\mathcal{B} based]}(\Gamma; \boldsymbol\omega')
&=\sum_{i=1}^n\sum_{j=1}^k u_{ij}\omega'_i\|\boldsymbol{\zeta}_i - \boldsymbol{\mu_\omega'}_j\|^2 \\
\\&=\sum_{j=1}^k \frac{1}{2{\Vol'}_j} \left(
2F_j{\Vol'}_j
-\left(\sum_{i \in C_j}  \sum_{\substack{\ell \in C_j}} 2s'_{i\ell} \right)
\right)  
\;  
\\&=\sum_{j=1}^k \frac{1}{2{\Vol'}_j} \sum_{i \in C_j} 
\sum_{\substack{\ell \in C_j}}
\left(\frac{{d'}_{ii}}{{d'}_{\ell\ell}}+\frac{{d'}_{\ell\ell}}{{d'}_{ii}}-2s'_{i\ell}\right) \label{eq:NRcmpwithGlove}
\;  
\\&=\sum_{j=1}^k F_j+  \sum_{j=1}^k\frac{1}{{\Vol'}_j} \left(
-\left(\sum_{i \in C_j}  \sum_{\substack{\ell \in C_j}} s'_{i\ell} \right)
\right)  
\;  
\end{align}
For $i \in C_j$
\begin{equation}
    d_{ii}-\sum_{\substack{\ell \in C_j\\ \ell\ne i}} s_{i\ell} 
    = \sum_{\substack{\ell \not\in C_j}} s_{i\ell} 
\end{equation}
Hence
\begin{equation}
    d'_{ii}-\sum_{\substack{\ell \in C_j}} s'_{i\ell} 
    = \sum_{\substack{\ell \not\in C_j}} s_{i\ell} 
\end{equation}
\begin{equation}
   \sum_{i \in C_j}  d'_{ii}-\sum_{i \in C_j} \sum_{\substack{\ell \in C_j}} s'_{i\ell} 
    = \sum_{i \in C_j}  \sum_{\substack{\ell \not\in C_j}} s_{i\ell} 
\end{equation}
\begin{equation}
   {\Vol'}_j-\sum_{i \in C_j} \sum_{\substack{\ell \in C_j}} s'_{i\ell} 
    = \sum_{i \in C_j}  \sum_{\substack{\ell \not\in C_j}} s_{i\ell} 
\end{equation}

That is
\begin{align} \label{eq:Q::NRkmeansW2}
Q^{[\mathcal{B} based]}(\Gamma; \boldsymbol\omega')
&=F+  \sum_{j=1}^k\frac{1}{{\Vol'}_j} \left(-{\Vol'}_j+
{\Vol'}_j
-\left(\sum_{i \in C_j}  \sum_{\substack{\ell \in C_j}} s'_{i\ell} \right)
\right)  
\;  
\\&=F+  \sum_{j=1}^k\frac{1}{{\Vol'}_j} \left(-{\Vol'}_j+
\sum_{i \in C_j}  \sum_{\substack{\ell \not\in C_j}} s_{i\ell}\right)
\;  
\\&=F-k+  \sum_{j=1}^k\frac{1}{{\Vol'}_j} 
\sum_{i \in C_j}  \sum_{\substack{\ell \not\in C_j}} s_{i\ell}
\;  
\;  \label{eq:QnKfin}
\end{align} 
(see \eqref{eq:qNRCut_def} for comparison). 

%

\section{Relationship Between \texorpdfstring{$\mathfrak{L_R}$}{LR}-based Clustering and 
\texorpdfstring{$\mathcal{B}$}{B}-based Clustering}\label{sec:NRLeqB}

From equations \eqref{eq:qNRCut_def} and  \eqref{eq:QnKfin} we see immediately that 
\begin{equation}
Q^{[\mathcal{B} based]}(\Gamma; \boldsymbol{\omega'})
= F-k+
Q^{[NRCut]}(\Gamma)
\end{equation}
As $F-k$ is a constant, minimizing one criterion minimizes the second one. 
As 
$\mathfrak{L_R}$-based Clustering 
approximates NRCut clustering, we see that we have here  
 a better situation than for $K$-embedding versus $L$-based clustering.  

\Bem{
\footnote{
Lingoes correction is needed, if $K$ turns out to have negative eigenvalues, see \cite{RAKMAKSTW:2020:trick}. 
The correction consists in adding $2\sigma$ to all elements of dissimilarity matrix $A$ except for the main diagonal, which has to stay equal to zero, where $\sigma\ge -\lambda_m$  where $\lambda_m$ is the smallest eigenvalue of $K$. Via adding we get a new matrix $A'$, for which we compute new $K'$ and use the prescribed embedding resulting from $K'$ and not from $K$, when performing $k$-means.  
}
}

\section{Equivalence Between \texorpdfstring{$\mathcal{B}$}{B}-based Clustering and weighted GloVe Vector Based Clustering}\label{sec:BeqwGloVe}

Consider the distance between two weighted vectors in GloVe vector space. 

Consider the squared distance between the document $i$ and $\ell$ in this space with $\omega`$ weights.   
Denoting ${d'}_{\ell\ell}=\omega_\ell' $
\begin{align}
\|g_\omega(\dok_i)-g_\omega(\dok_\ell)\|^2 =&
\|g_\omega(\dok_i)\|^2+ g_\omega(\dok_\ell)\|^2
- 2g_\omega(\dok_i)^Tg_\omega(\dok_\ell)
\\ =& \frac{1}{{d'}_{ii}^2} + \frac{1}{{d'}_{\ell\ell}^2} -2 \frac{s_{i\ell}}{{d'}_{ii}{d'}_{\ell\ell}}
\end{align}

Under the previously assumed notation, we can write the target function of the 
weighted GloVe Vector Based Clustering:

\begin{align}  
Q^{[{\omega'}GloVe ]}(\Gamma; \boldsymbol\omega')
=&\sum_{j=1}^k \sum_{i\in C_j} \omega'_i\|g_\omega(\dok_i)-\boldsymbol\mu_{\boldsymbol{\omega'}}(C_j)\|^2 
\\=&\sum_{j=1}^k \frac{1}{{2\Vol'}_j} \sum_{i\in C_j} \sum_{\ell\in C_j}
\omega'_i \omega'_\ell
\|g_\omega(\dok_i)-g_\omega(\dok_\ell)\|^2  
\\=&\sum_{j=1}^k \frac{1}{{2\Vol'}_j} \sum_{i\in C_j} \sum_{\ell\in C_j}
\omega'_i \omega'_\ell
\left(\frac{1}{{d'}_{ii}^2} + \frac{1}{{d'}_{\ell\ell}^2} -2 \frac{s_{i\ell}}{{d'}_{ii}{d'}_{\ell\ell}}  \right)
\\=&\sum_{j=1}^k \frac{1}{{2\Vol'}_j} \sum_{i\in C_j} \sum_{\ell\in C_j}
\left(\frac{{d'}_{\ell\ell}}{{d'}_{ii}} + \frac{{d'}_{ii}}{{d'}_{\ell\ell}} -2 s_{i\ell}  \right)
\end{align}
This is identical with eq. \eqref{eq:NRcmpwithGlove}. 
This means one can use $N$-based or $R$-based GSC methods methods for similarities derived from GloVe embeddings (instead of clustering in higher dimensional GloVe space) and still be able to provide cluster membership explanations related to GloVe embeddings.

\section{Experiments}
\label{sec:experiments}
We performed a series of experiments to find out how well the GSC deals with documents similarity of which was identified based on GloVe embedding and what effects we get by explanation approach presented here.

Three aspects are of particular interest.   {First, the question is whether GloVe-like embedding contributes to better extraction of internal clusters in the data.}
 {Second, does an explanation based on GloVe embedding, which is computationally more expensive than TVS embedding, provide a new quality. }  {And finally, we ask whether only some specific GSC methods gain some advantage, or whether the phenomenon is broader, for other types of GSCs.} 

In our experiments, 
we used two GloVe embeddings\footnote{Available from the Web Page \url{https://nlp.stanford.edu/projects/glove/}; for a description see \cite{Pennington:2014:glove}.}:
\begin{itemize}
    \item glove.6B.100dm trained on Wikipedia data, 
    taken from 
    \url{https://nlp.stanford.edu/data/glove.6B.zip},
    with 400,000 words/tokens in 100-dimensional space,  
    to which we refer later as WikiGloVe embedding
    \item glove.Twitter.27B.100d, trained specially for the Twitter domain, taken from 
    \url{https://nlp.stanford.edu/data/glove.twitter.27B.zip}, 
    with 1,193,514 words/tokens in 100-dimensional space, 
    to which we refer later as TweetGloVe embedding
    \footnote{
     {Although both embeddings are huge in terms of words, their actual content in terms of real English words is limited. }
    The so-called american-english-insane.ascii dictionary contains 507871.  Their intersection with TweetGloVe dictionary yields only 72821 words, 
    with WikiGloVe - only 104636 words. The remaining content appears to be of a low quality.  
    }
\end{itemize}

For the experiments we used the following real data collections:

\begin{table}[ht]
\centering
\begin{tabular}{|r|l|c|}
\hline
    No.  & hashtag & count \\
    \hline
  0& 90dayfiance & 316\\
	 1& tejran & 345\\
	 2& ukraine & 352\\
	 3& tejasswiprakash & 372\\
	 4& nowplaying & 439\\
  \hline
\end{tabular}
\begin{tabular}{|r|l|c|}
\hline
    No.  & hashtag & count \\
    \hline
	 5& anjisalvacion & 732\\
	 6& puredoctrinesofchrist & 831\\
	 7& 1 & 1105\\
	 8& lolinginlove & 1258\\
	 9& bbnaija & 1405\\
  \hline
\end{tabular}
\caption{TWT.10 dataset - hashtags and cardinalities of the set of related tweets  used in the experiments}\label{tab:twt10set}
\end{table}


\begin{itemize}
  \item TWT.10 -- the set of random tweets published on Twitter (currently X) between 2019 and 2023, 
    of length greater than 150 characters each, containing exactly one of the hashtags listed in Table~\ref{tab:twt10set}.\footnote{The dataset will be made available upon publication.}
  \item TWT.3 -- a subset of TWT.10 consisting of tweets containing the hashtags: 
    \tht{anjisalvacion}, \tht{nowplaying} and \tht{puredoctrinesofchrist}. 
\end{itemize}

We will present experimental work on three aspects
\begin{itemize}
    \item visualization of clustering behavior under GloVe embedding (Section \ref{sec:clusterVisual}),
    \item intuitiveness of explanations (Section \ref{sec:clusterExplain}),
    \item clustering behavior for various variants of GSC (Section \ref{sec:clusterGSCvariants}). 
\end{itemize}

\subsection{Visualization of Clustering Behavior}
\label{sec:clusterVisual}

In this section, we will demonstrate how GloVe embeddings affect various GSC methods. We mainly use WikiGloVe embedding, and we compare the results with TweetGloVe and TVS embedding. 
We will demonstrate this with a collection of TWT.3 tweets containing exactly one of the three hashtags:  \tht{anjisalvacion}, 
\tht{puredoctrinesofchrist}, 
\tht{nowplaying} of cardinalities  732, 831, 439 resp. 

\begin{figure}
\begin{center}
\includegraphics[width=0.49\textwidth]{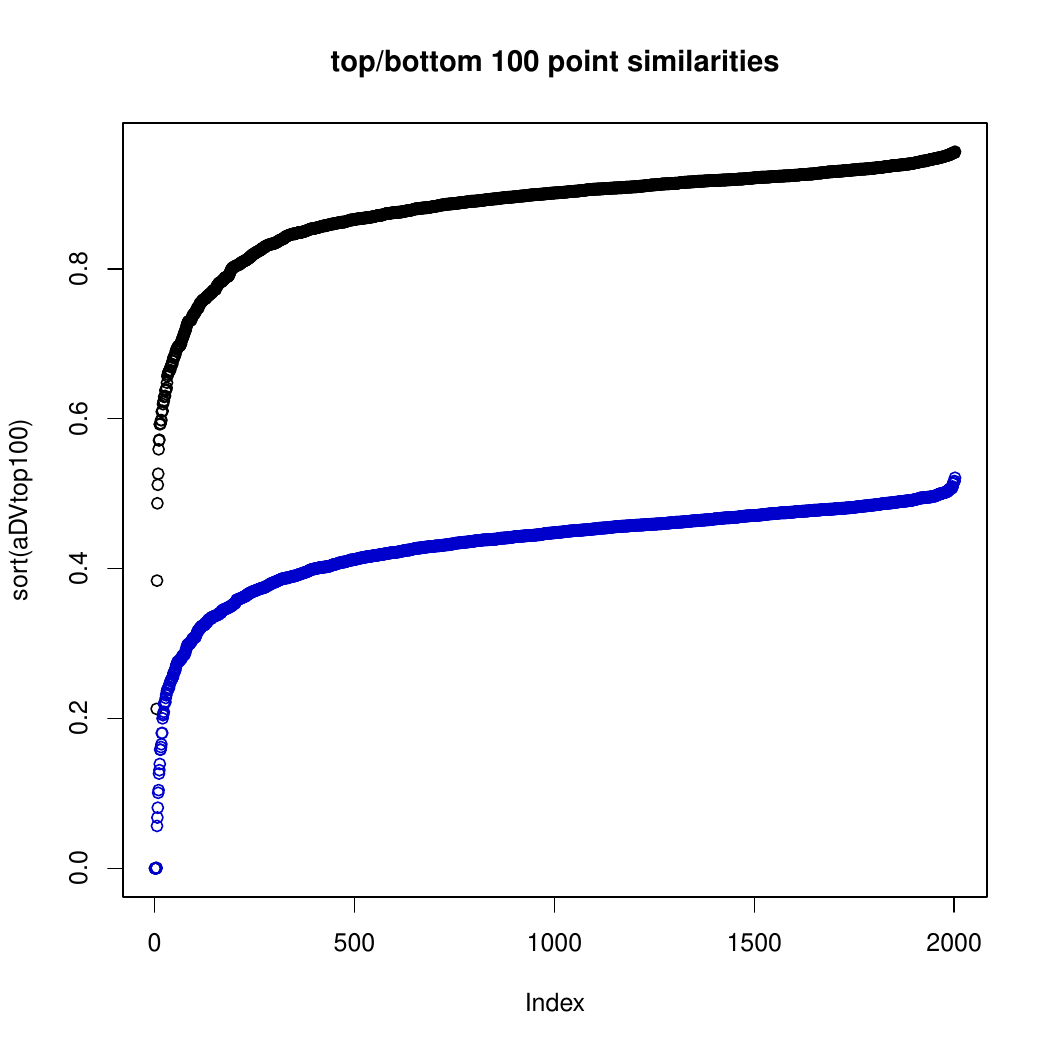} %
\includegraphics[width=0.49\textwidth]{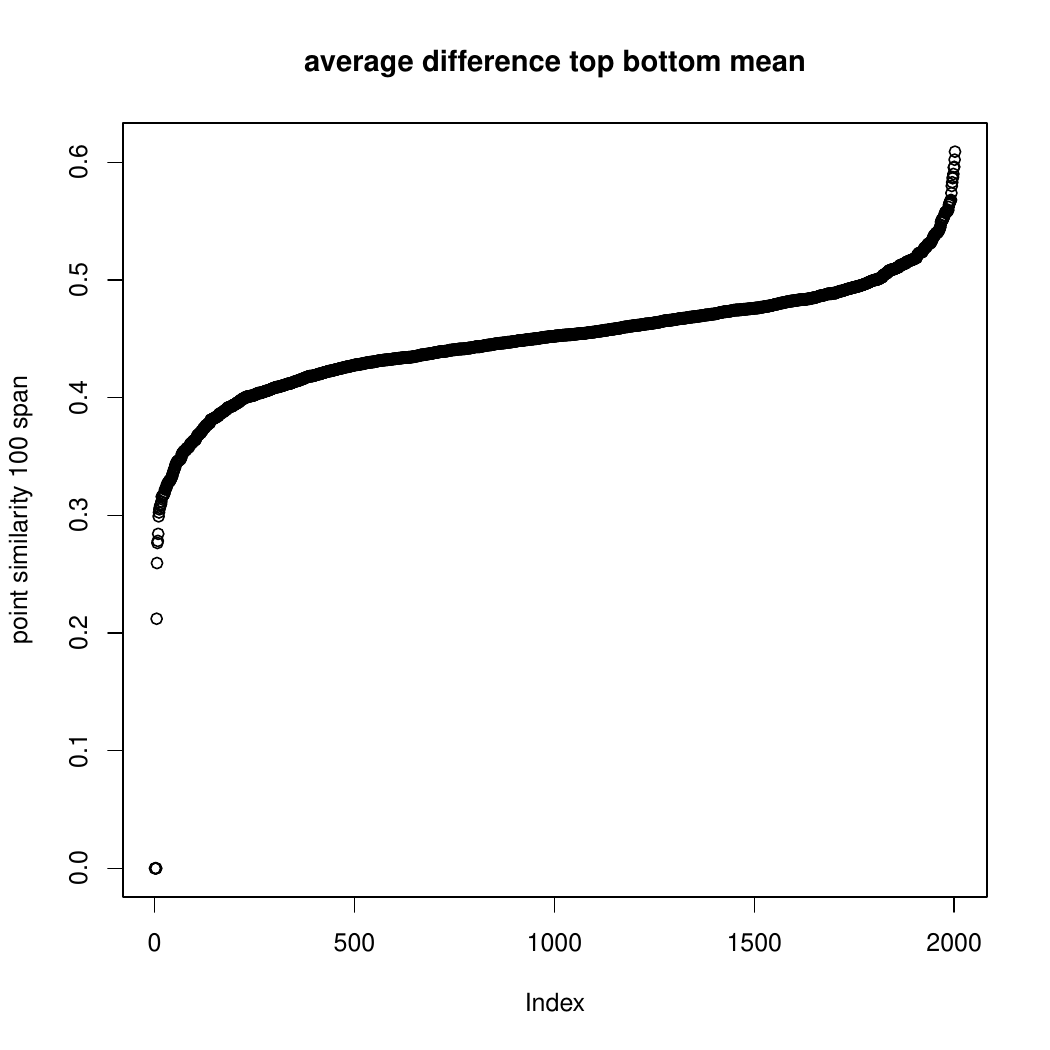}%
 \end{center}
\caption{Left: Objects sorted by increasing average similarity among its top 5\% similarities (black line) and by 
increasing average similarity among its lowest 5\% similarities (green line).
Right: 
Objects sorted by increasing difference between average similarity among its top 5\% similarities and  average similarity among its lowest 5\% similarities.
Dataset TWT.3.  Embedding WikiGloVe.
}\label{fig:TWT3ht_toplowsim}
\end{figure}

\begin{figure}
\begin{center}
\includegraphics[width=0.49\textwidth]{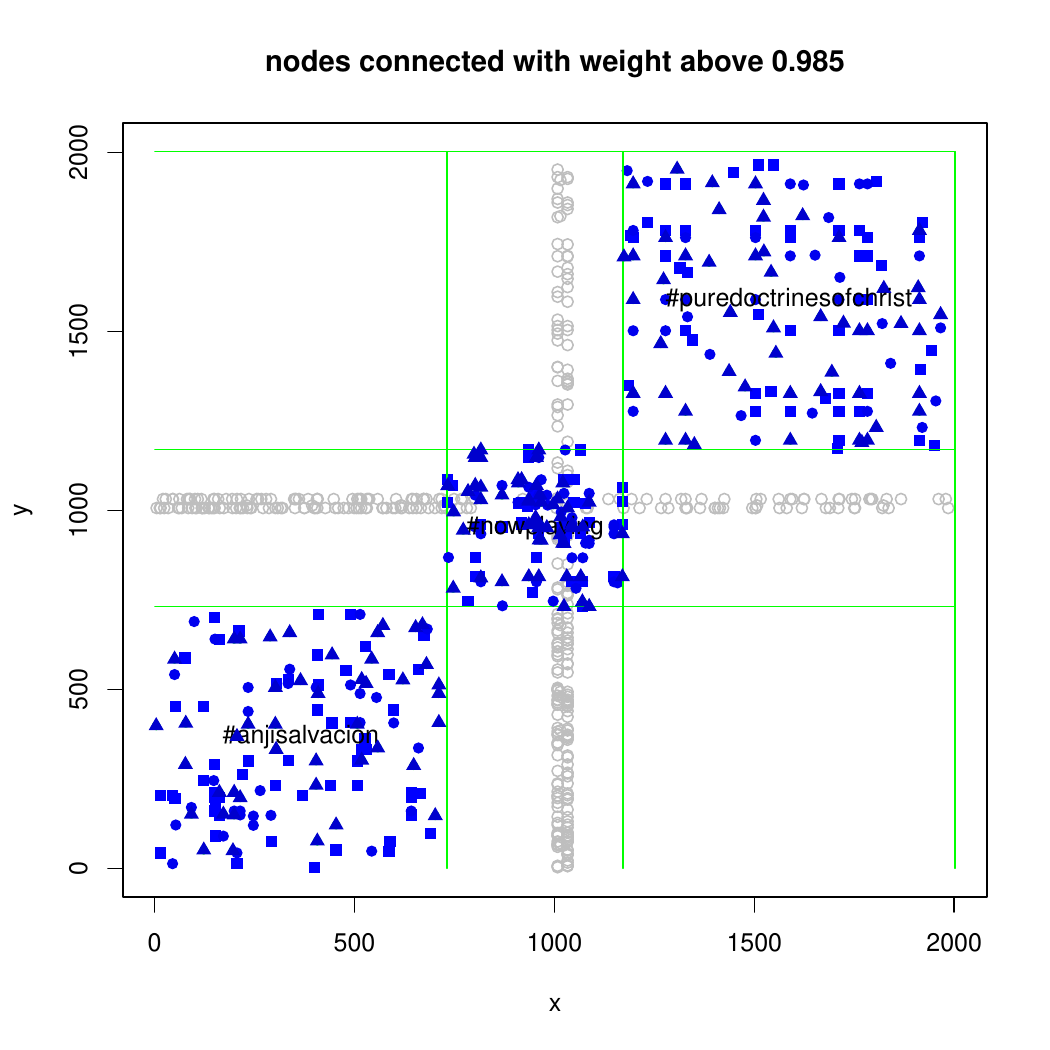} \newline%
\includegraphics[width=0.49\textwidth]{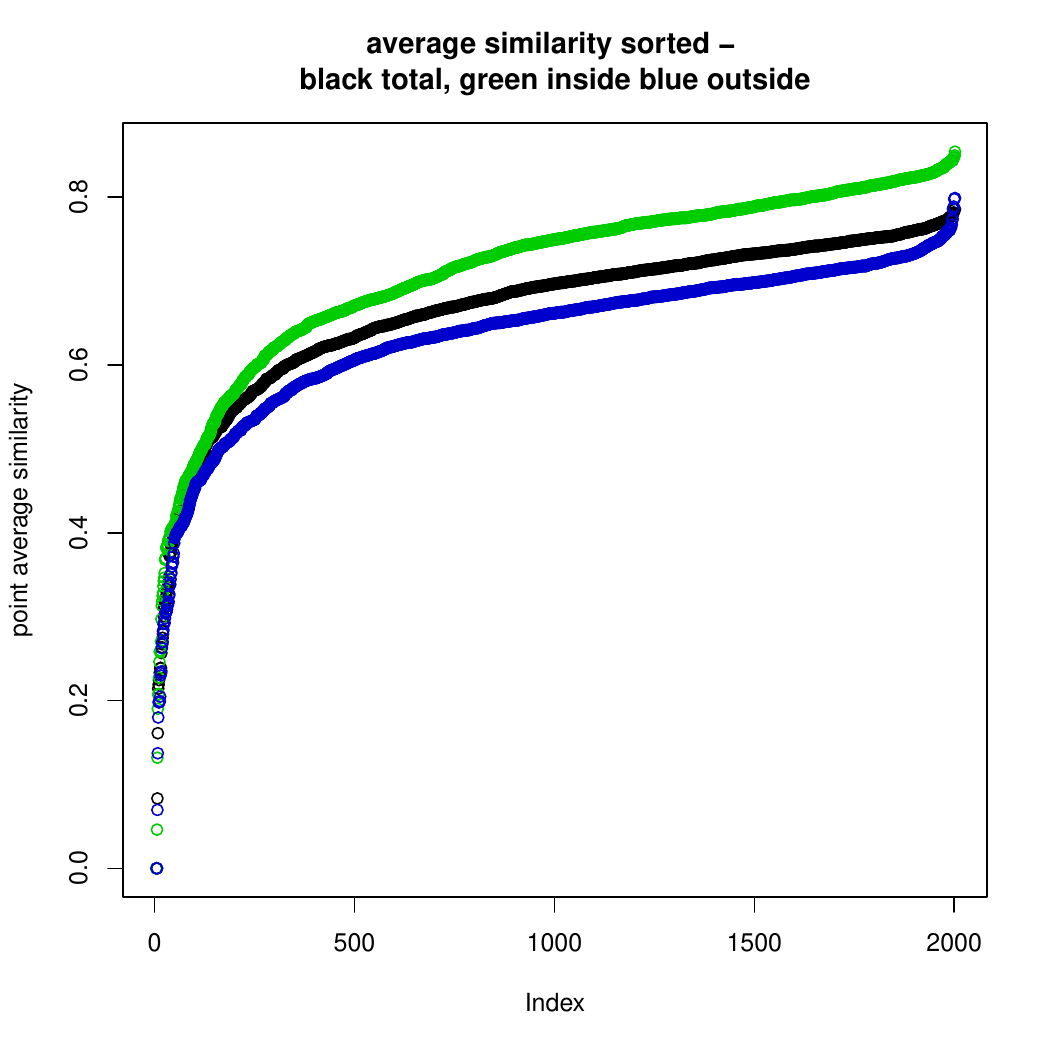} %
\includegraphics[width=0.49\textwidth]{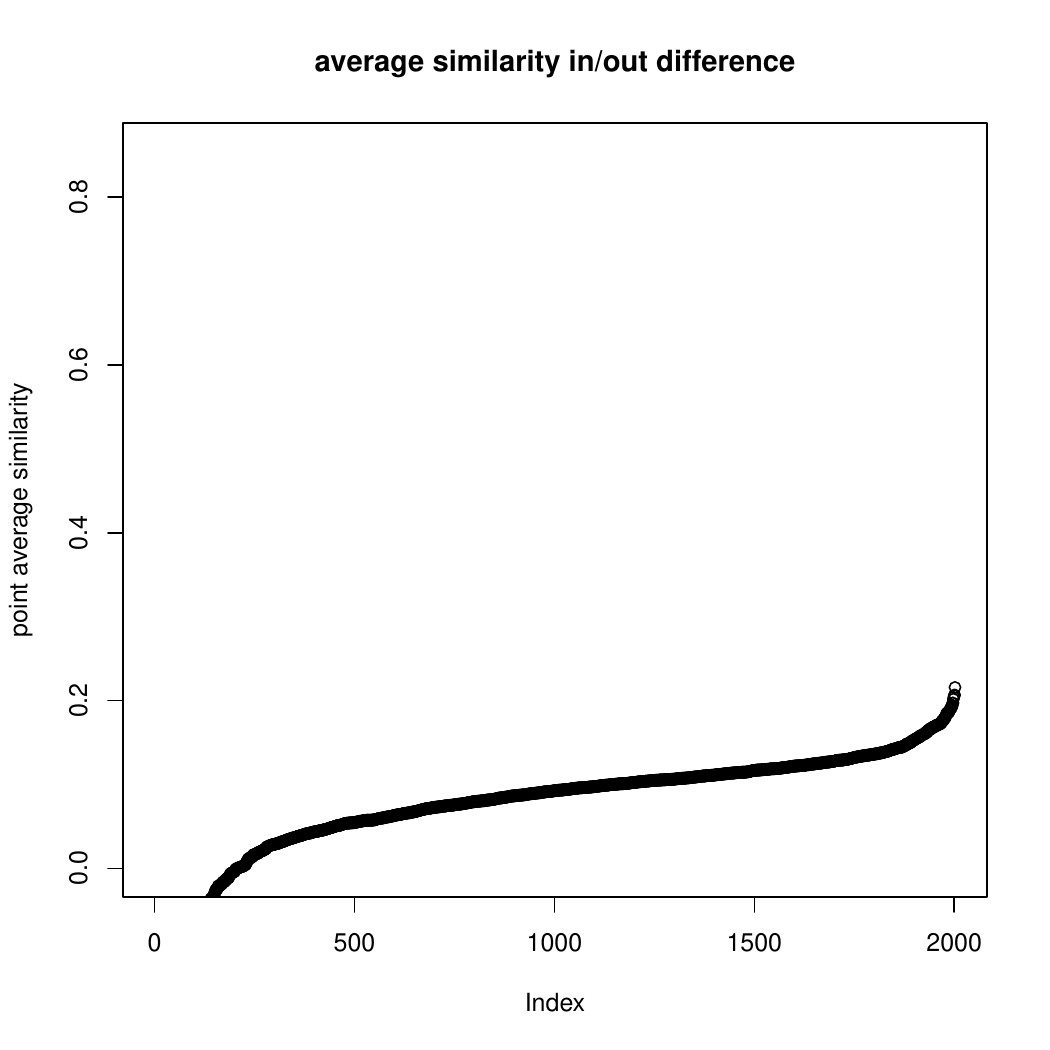} %
 \end{center}
\caption{
Top: Largest and lowest similarities within intrinsic clusters. Left: Average similarities within and outside of intrinsic clusters. Right: Differences between average similarities within and outside of clusters.  
Dataset TWT.3. Embedding WikiGloVe.
}\label{fig:TWT3ht_trueclusters}
\end{figure}

Figure \ref{fig:TWT3ht_toplowsim} presents an overview of the basic properties of the TWT.3 dataset, while Figure \ref{fig:TWT3ht_trueclusters} shows its properties with respect to intrinsic clustering (that is with respect to hashtags attached by humans). 
One sees from Figure \ref{fig:TWT3ht_toplowsim} that for WikiGloVe embedding there exists a considerable span among similarities of any document to other documents and that there exist documents much less similar to a document than some other. This should be helpful when identifying clusters, though the right figure suggests that there exist a couple of documents where this difference is negligible, so their clustering may be poor.   

\begin{figure}
\begin{center}
\includegraphics[width=0.49\textwidth]{TWT3ht_mxlinks.pdf} \newline%
\includegraphics[width=0.49\textwidth]{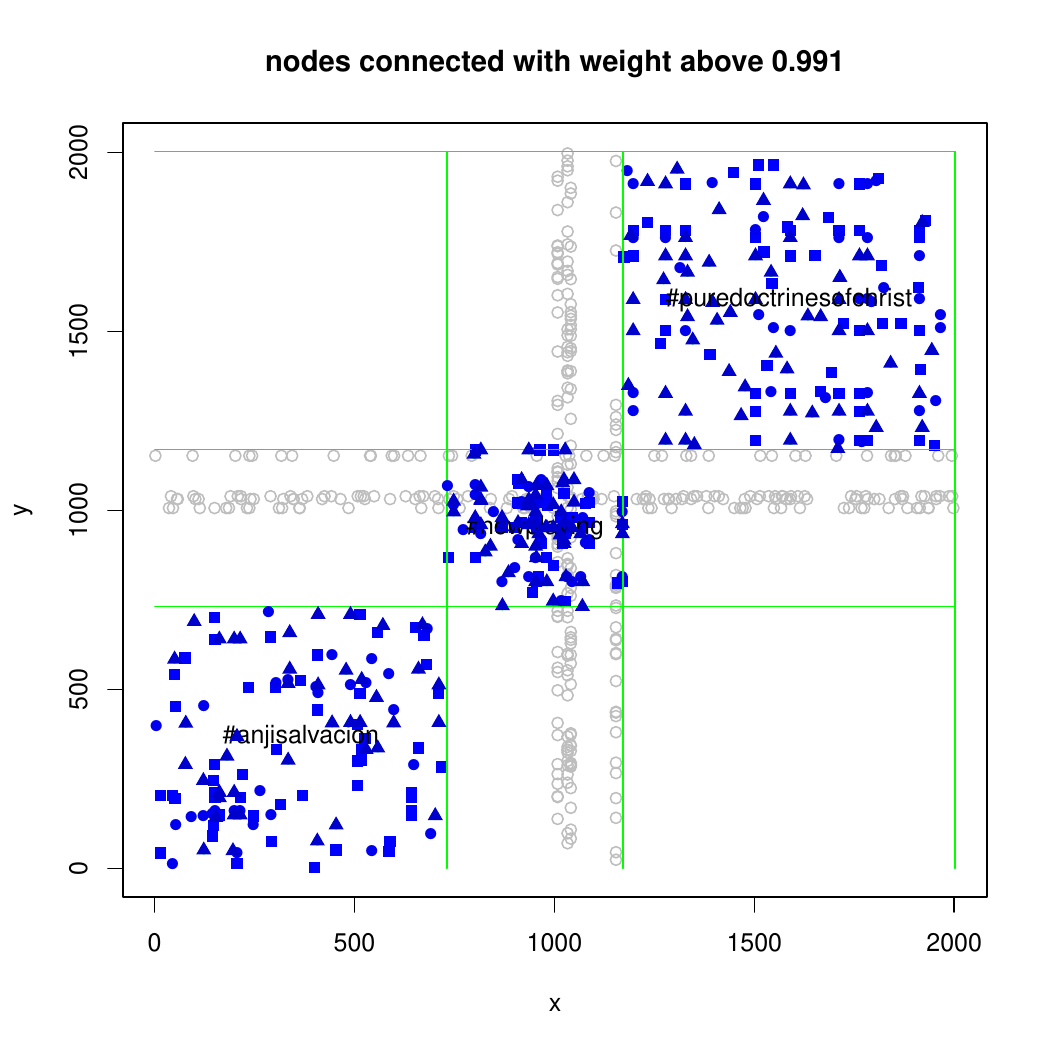} %
\includegraphics[width=0.49\textwidth]{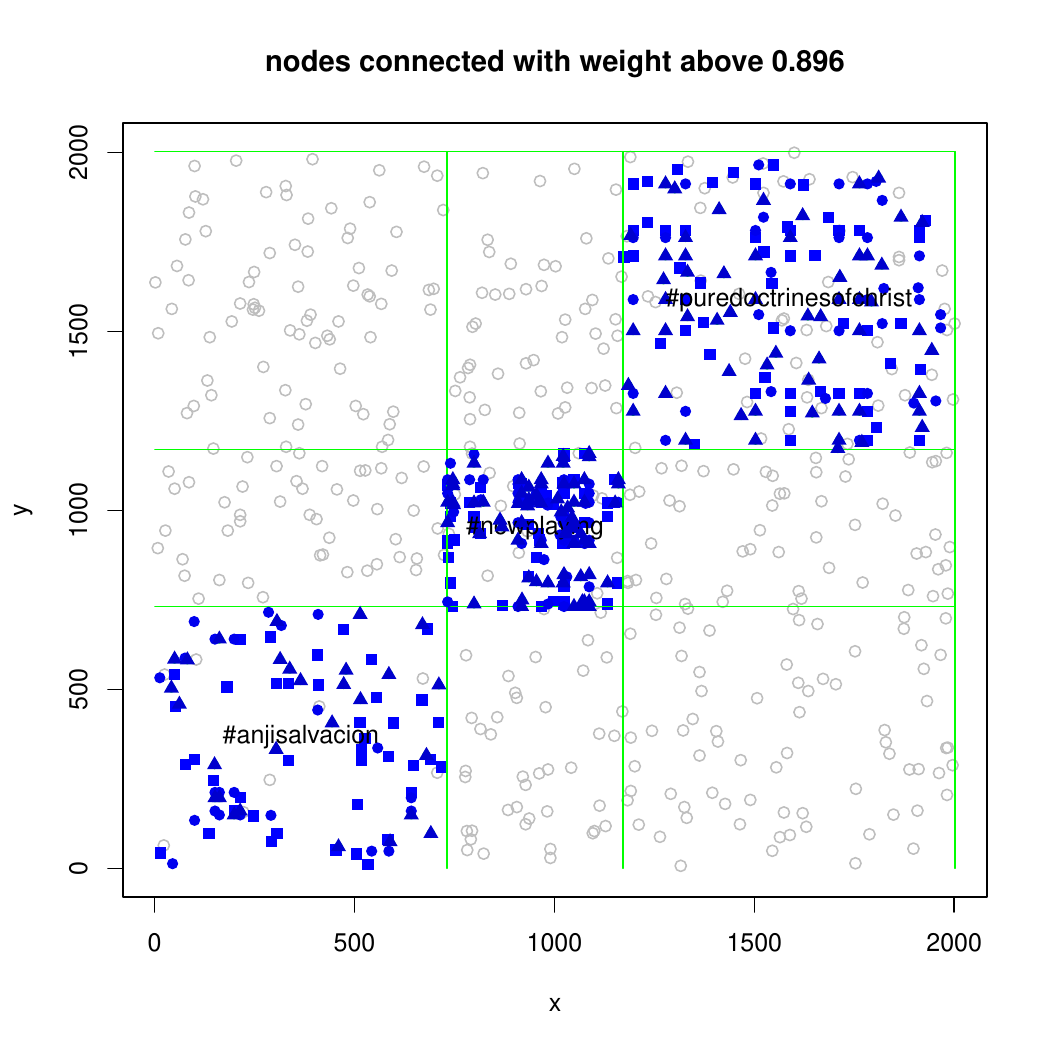} %
 \end{center}
\caption{
Largest and lowest similarities within intrinsic clusters under different embeddings: 
Top: WikiGloVe, Left: TweetGloVe, Right: TVS.
Dataset TWT.3. 
}\label{fig:TWT3ht_mxlinks_cmp}
\end{figure}

\begin{figure}
\begin{center}
\includegraphics[width=0.49\textwidth]{TWT3ht_siminoutdiff.pdf} %
\newline
\includegraphics[width=0.49\textwidth]{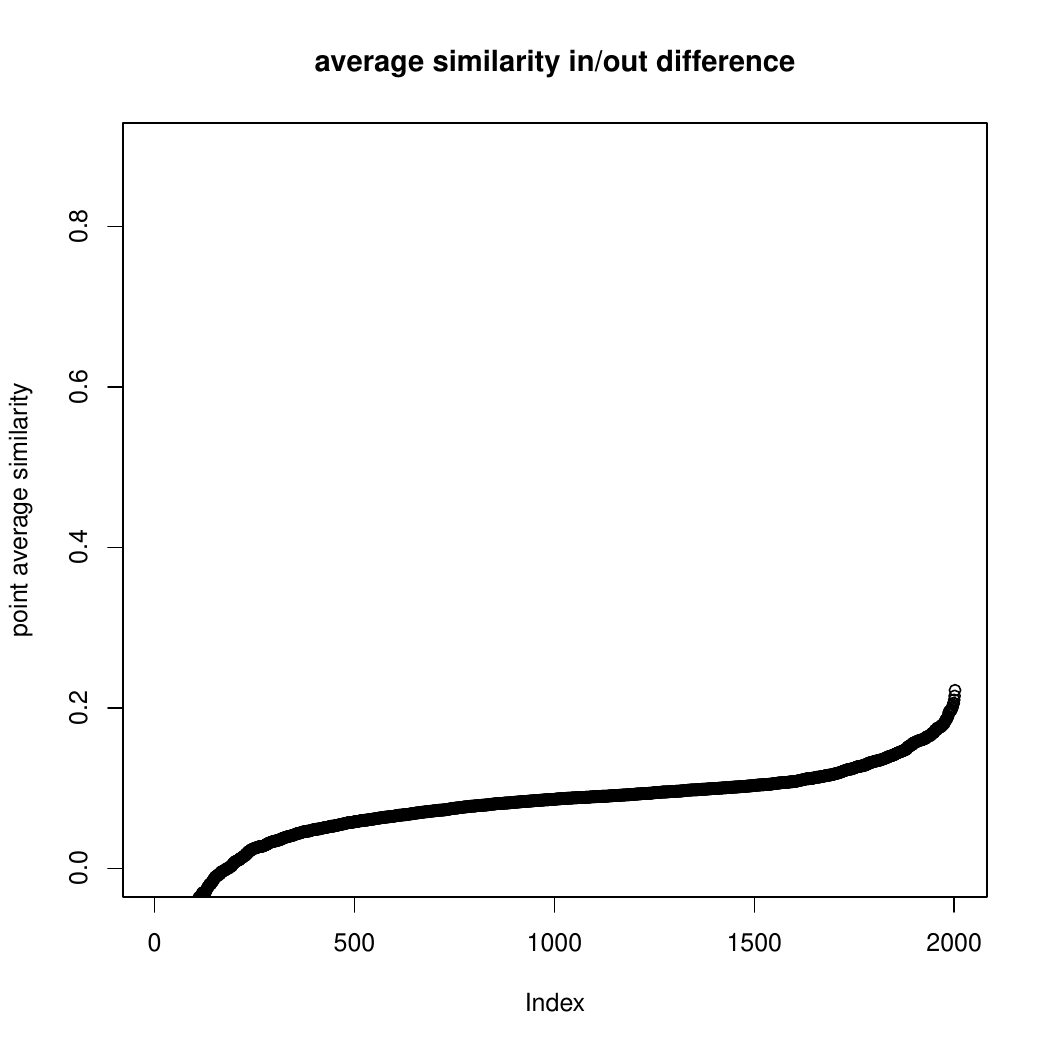} %
\includegraphics[width=0.49\textwidth]{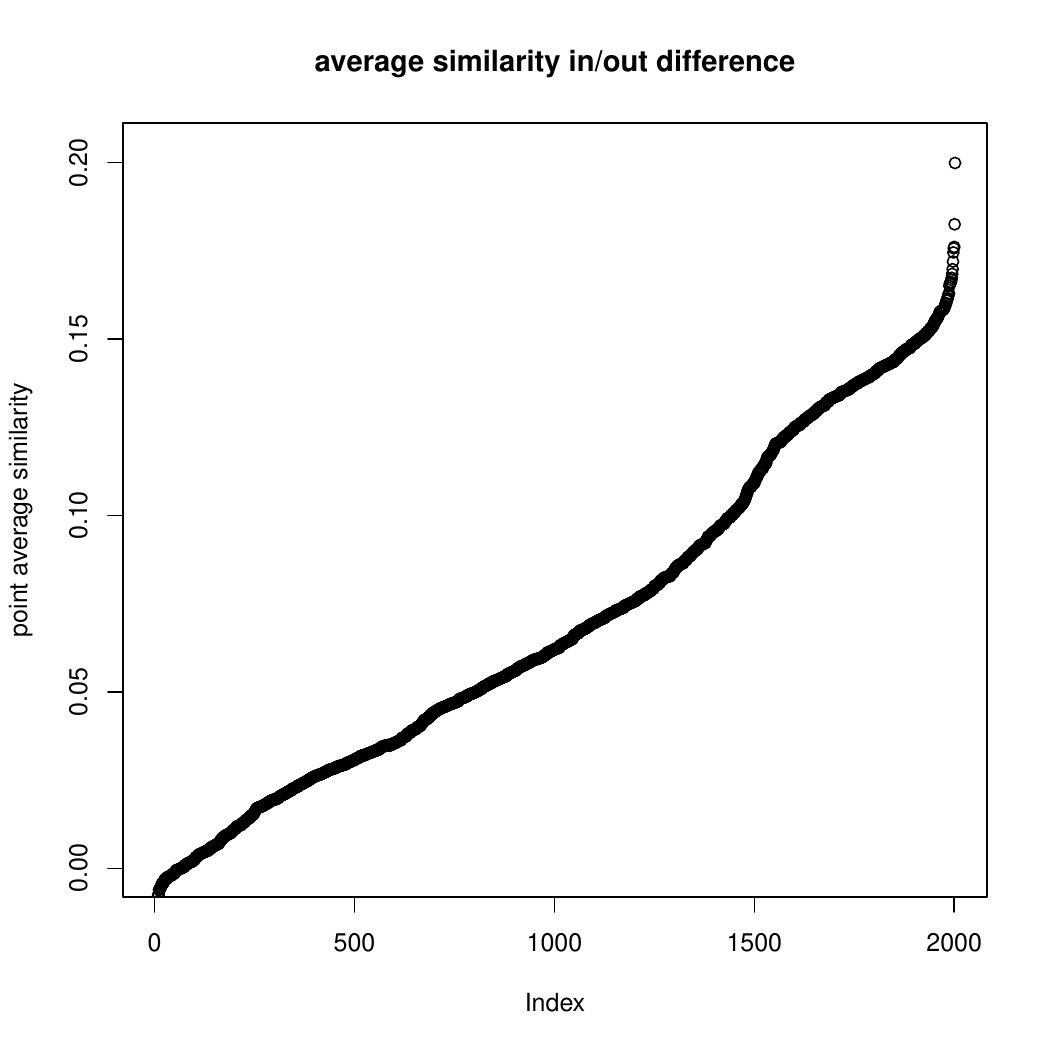} %
 \end{center}
\caption{
Differences between average similarities within and outside of clusters.  
Top: WikiGloVe, Left: TweetGloVe, Right: TVS.
Dataset TWT.3. 
}\label{fig:TWT3ht_siminoutdiff_cmp}
\end{figure}

The top picture in Figure 
\ref{fig:TWT3ht_trueclusters} reveals that the cluster \tht{nowplaying} has suspicious documents - the weakest similarities are concentrated in this cluster. 
A look at Figure \ref{fig:TWT3ht_mxlinks_cmp} reveals that the same problem exists under TweetGloVe embedding, while it is absent under TVS embedding. 
The right picture in Figure 
\ref{fig:TWT3ht_trueclusters} shows that some 100 documents are, on average, more similar to other clusters than to their own in WikiGloVe. 
We have the same problem under TweetGloVe, as visible in Figure \ref{fig:TWT3ht_siminoutdiff_cmp}, while it is not present in TVS embedding. However, under TVS a much larger number of documents has a quite low difference between these similarities.

\begin{figure}
\begin{center}
\includegraphics[width=0.3\textwidth]{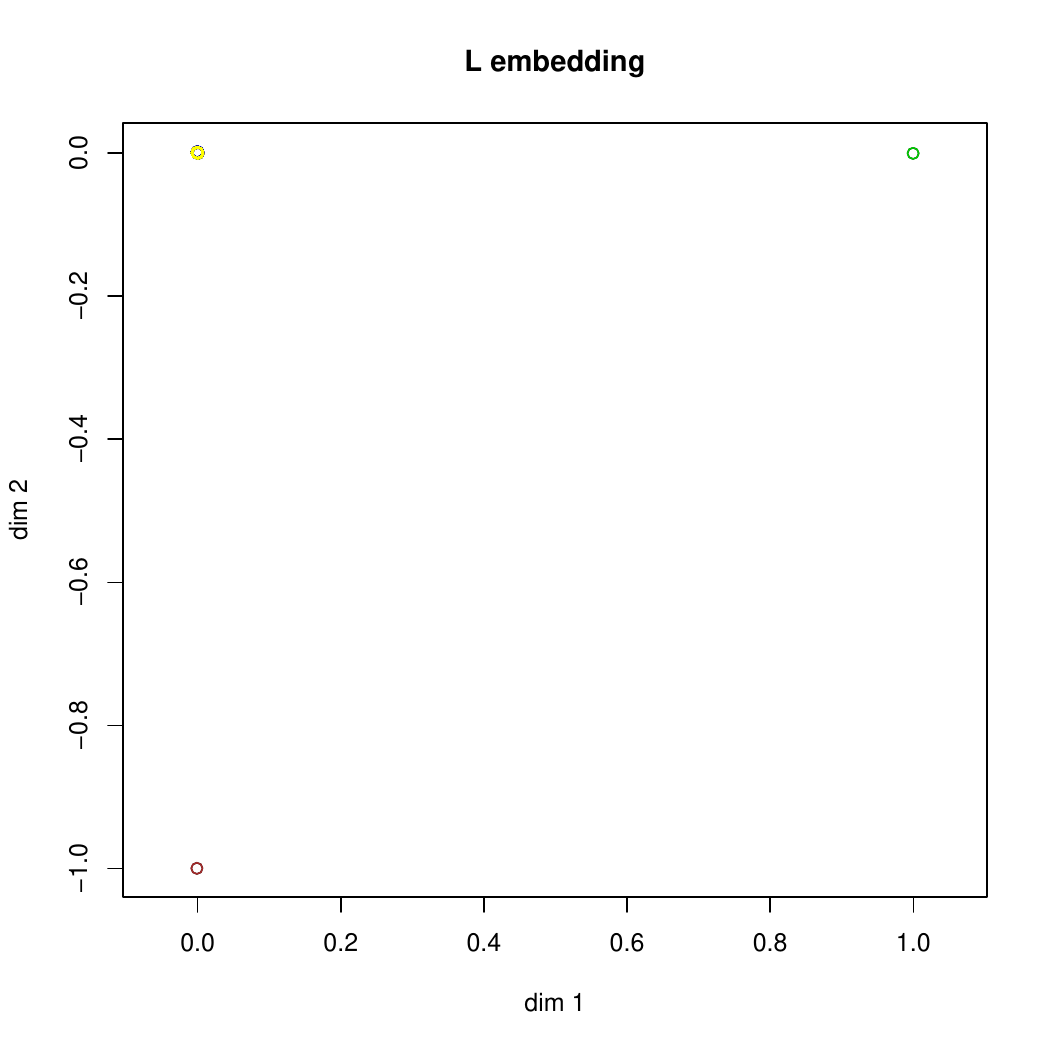} %
\includegraphics[width=0.3\textwidth]{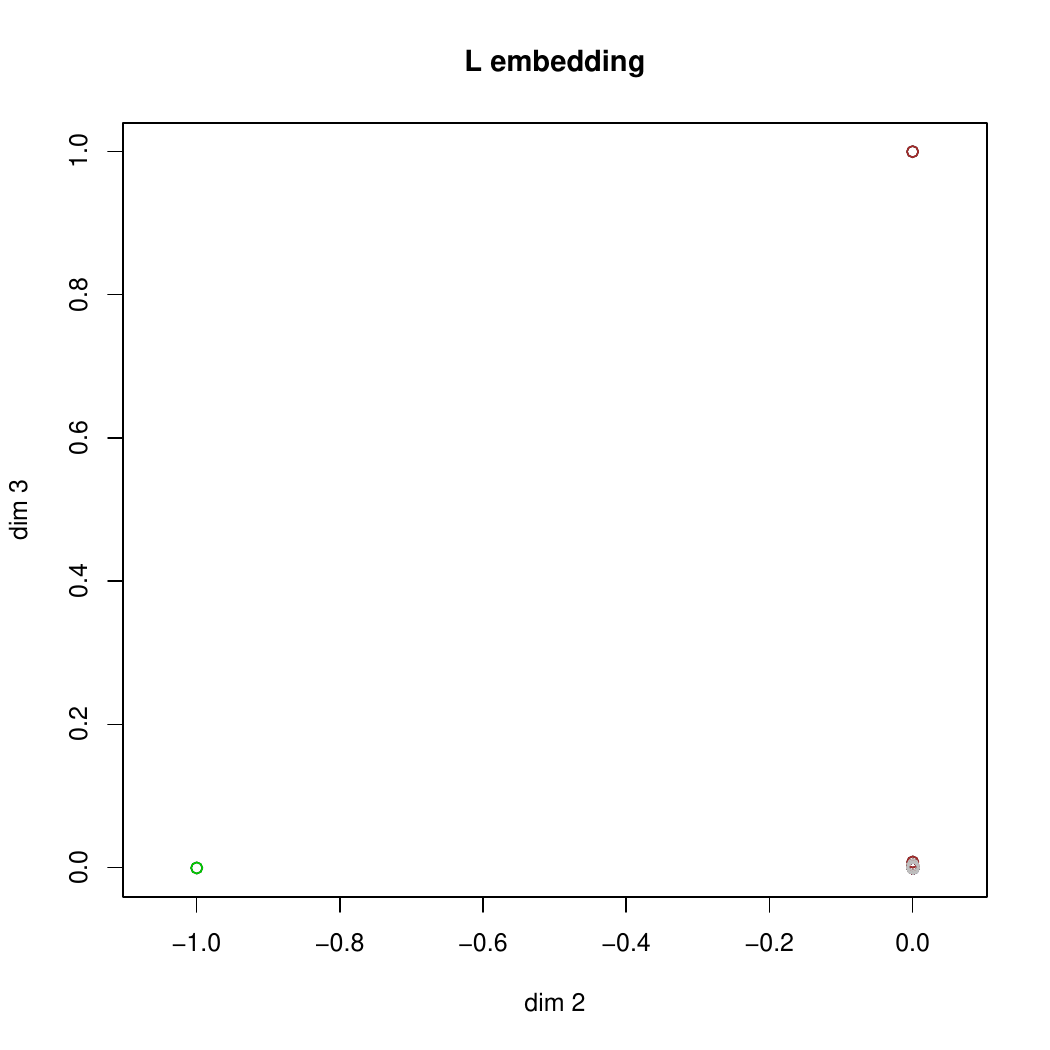} %
\includegraphics[width=0.3\textwidth]{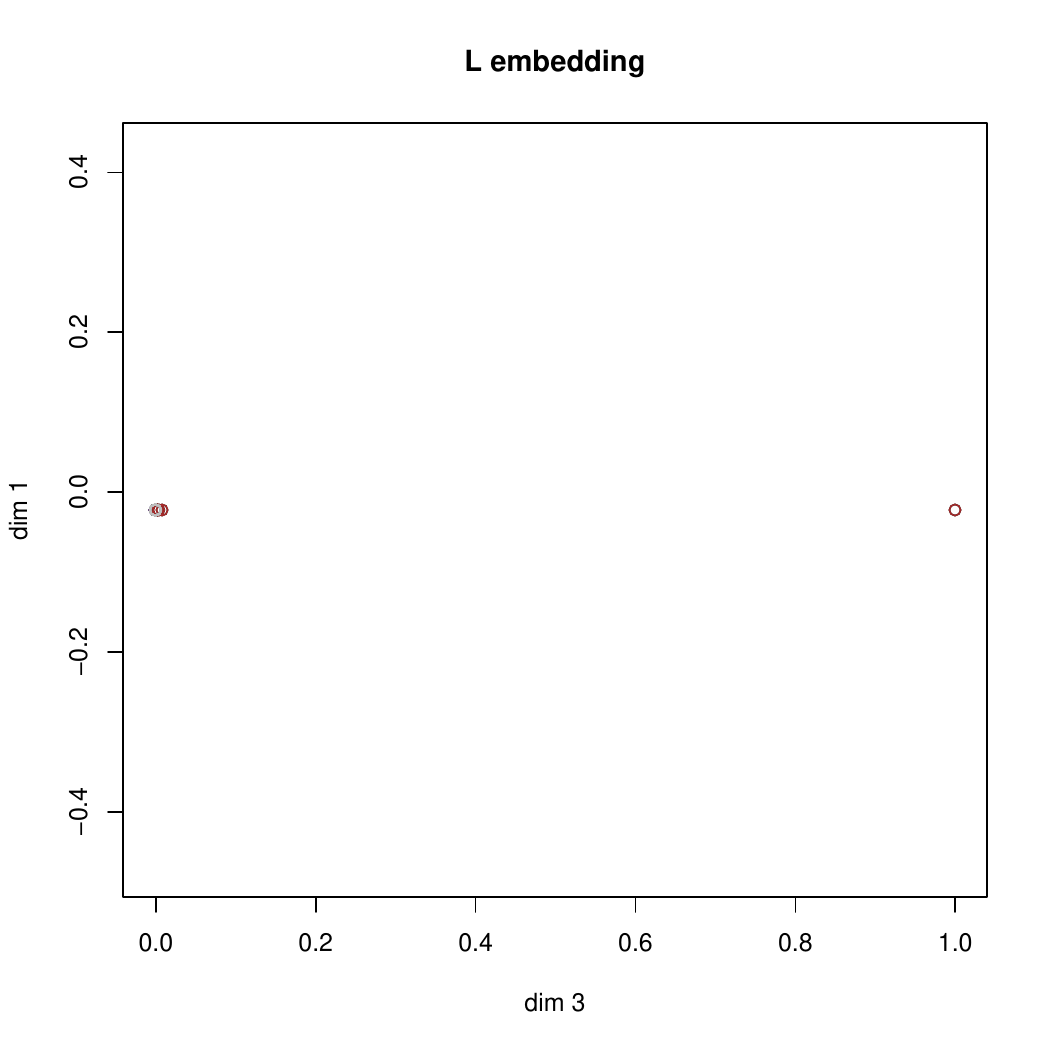} %
 \end{center}
\caption{A glance at the $L$-based clustering from three different perspectives (axes 1,2, or 2,3 or 1,3).  Different colors reflect the clusters.   Dataset TWT.3. Embedding WikiGloVe.
}\label{fig:TWT3ht_Lbased}
\end{figure}

Figure \ref{fig:TWT3ht_Lbased}
shows the clustering results using the $L$-based method. Table \ref{tab:trueL} provides with numerical information.
We observe that there are a few documents located quite far away from each other, resulting in small clusters and a larger one. Our experiments show that the very same problem is observed for the other embeddings (TweetGloVe and TVS).

\begin{figure}
\begin{center}
\includegraphics[width=0.3\textwidth]{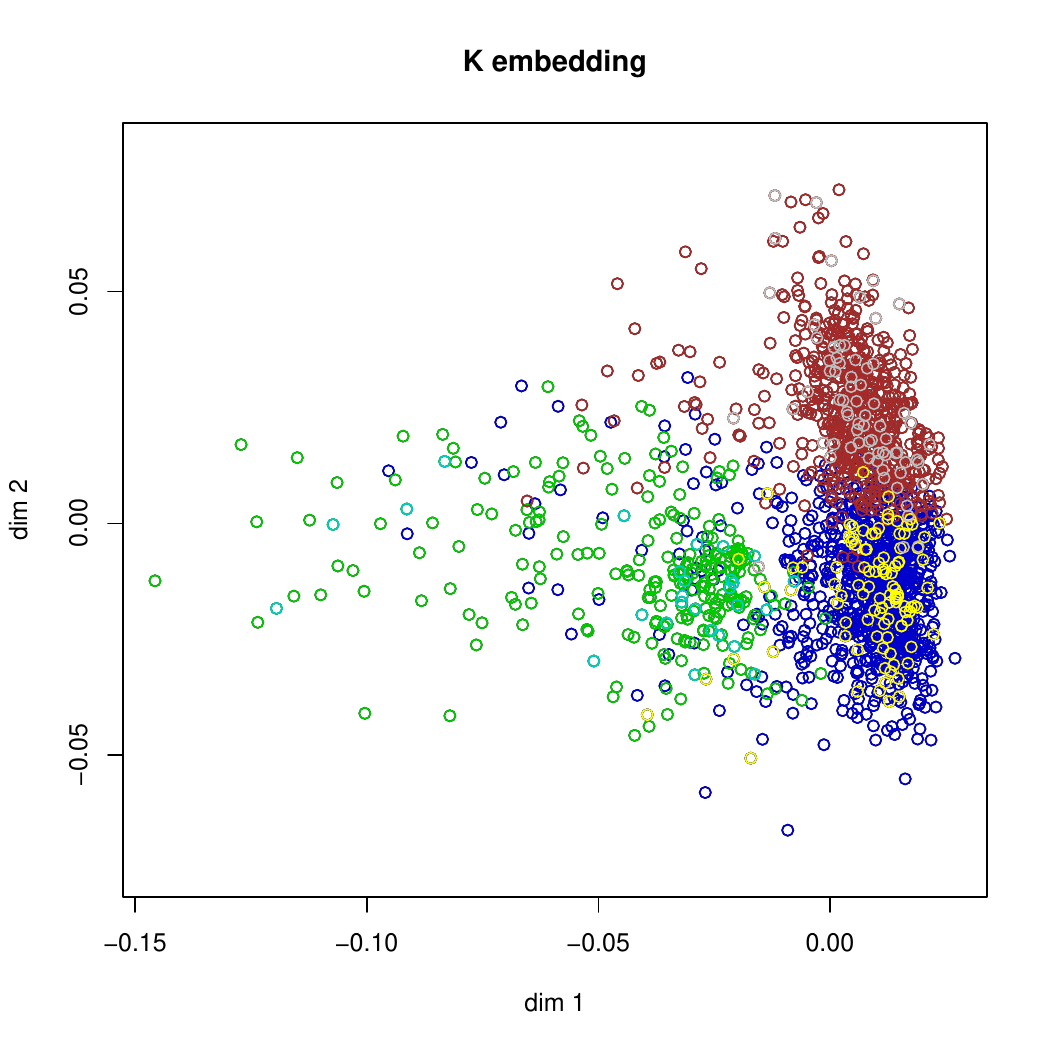} %
\includegraphics[width=0.3\textwidth]{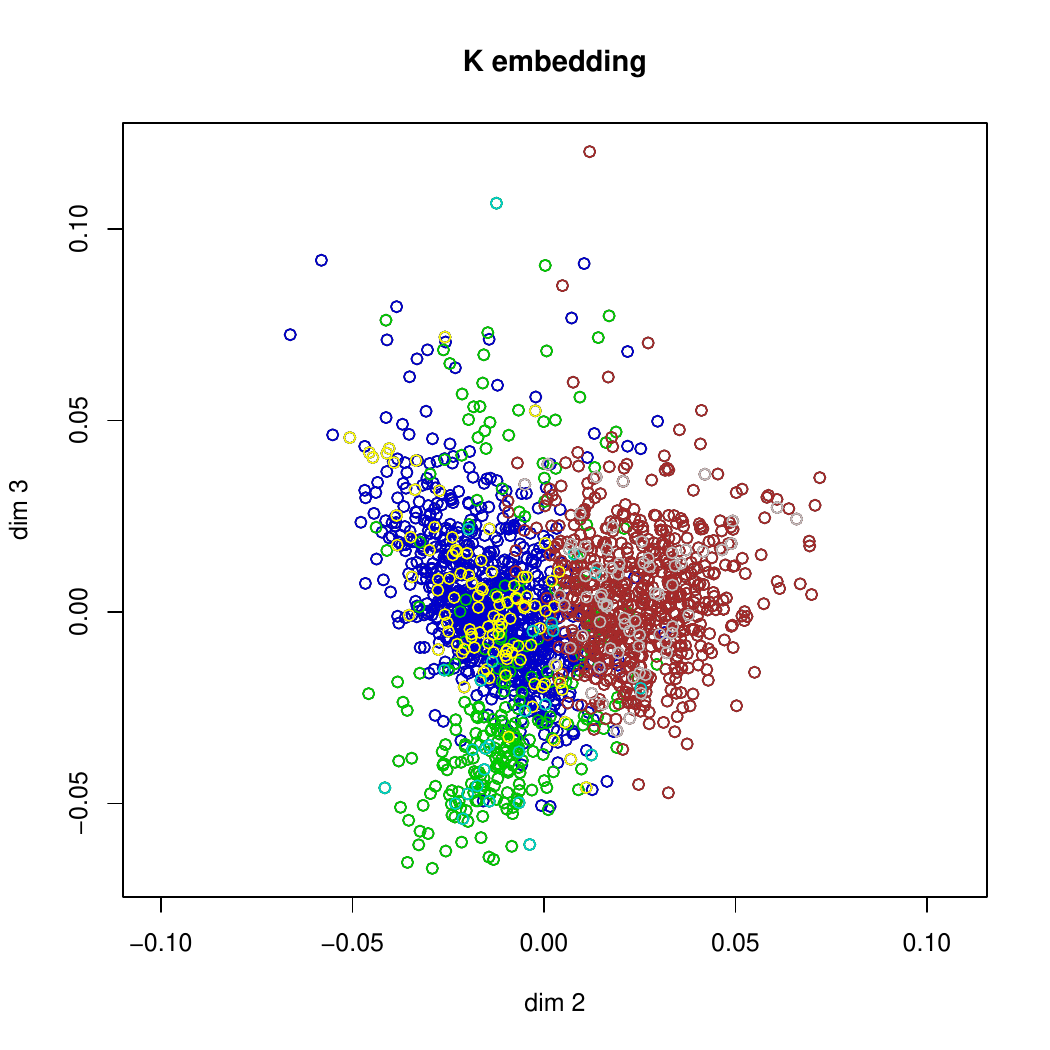} %
\includegraphics[width=0.3\textwidth]{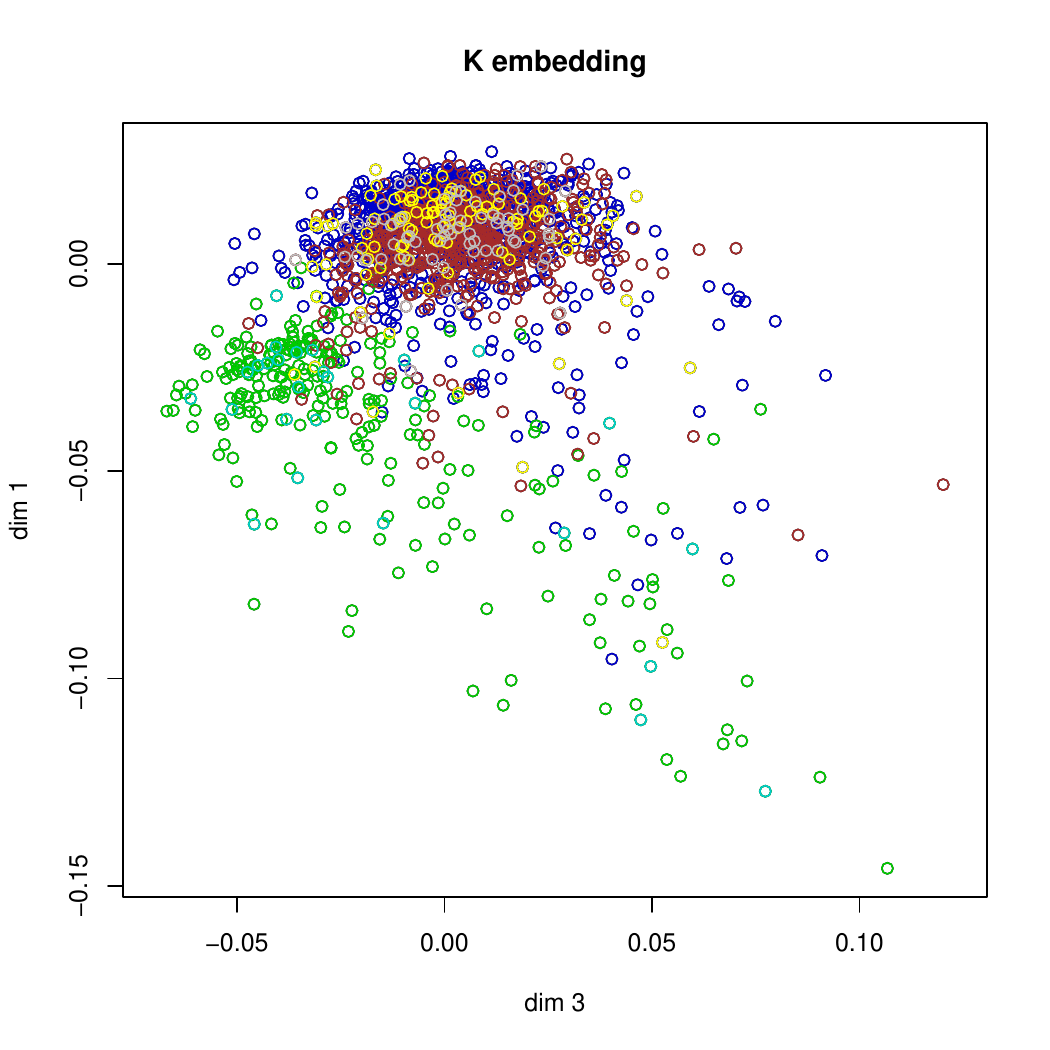} %
 \end{center}
\caption{A glance at the $K$-based clustering from three different perspectives (axes 1,2, or 2,3 or 1,3).  Different colors reflect the clusters.   Dataset TWT.3. Embedding WikiGloVe.
}\label{fig:TWT3ht_Kbased}
\end{figure}

Figure \ref{fig:TWT3ht_Kbased}
shows the clustering results using the $K$-based method. Table \ref{tab:trueK} provides with numerical information. Documents appear to fill the space in the core region, but some documents are scattered a bit further away from it, which looks much better than $L$-based clustering. This allows clusters to be detected near real clusters (hashtags).

\begin{figure}
\begin{center}
\includegraphics[width=0.3\textwidth]{TWT3ht_Kemb_12.pdf} %
\includegraphics[width=0.3\textwidth]{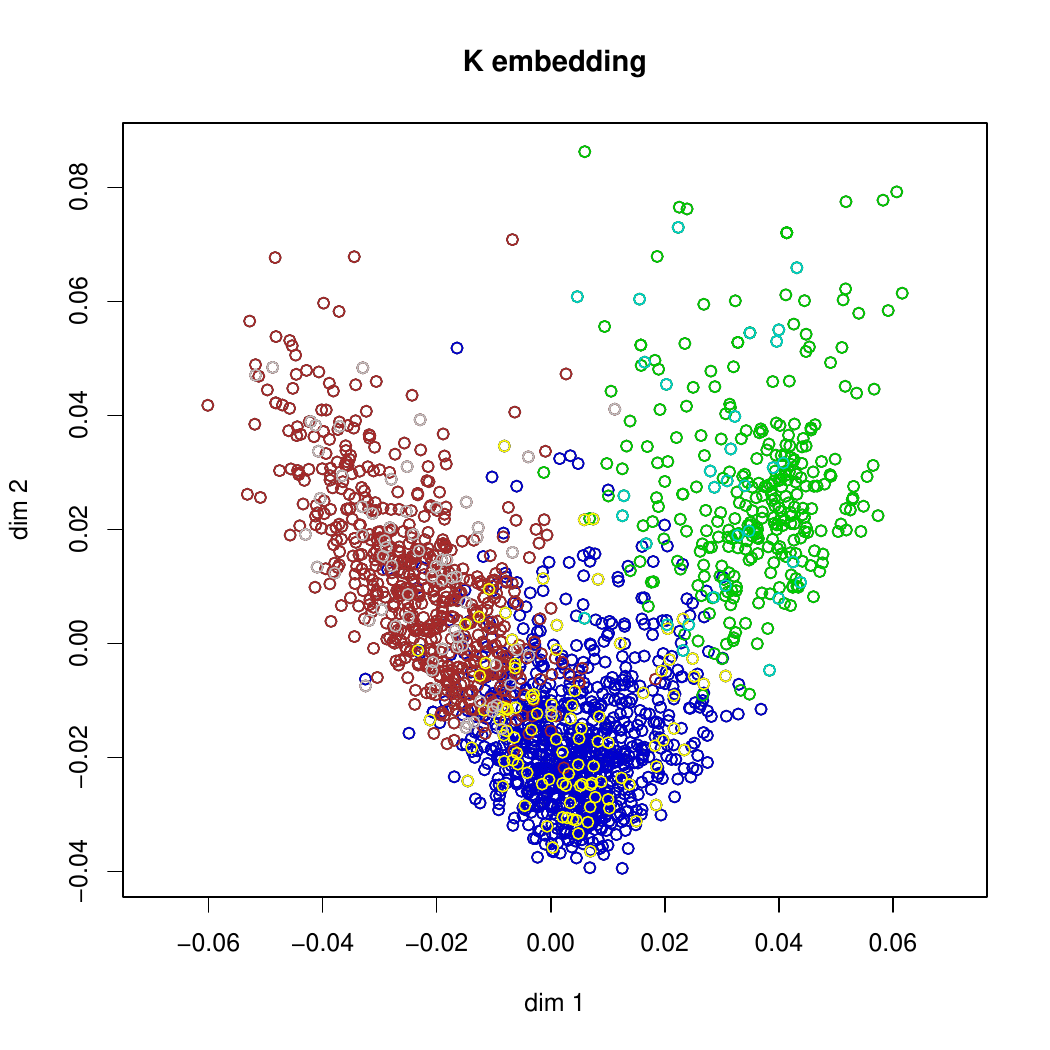} %
\includegraphics[width=0.3\textwidth]{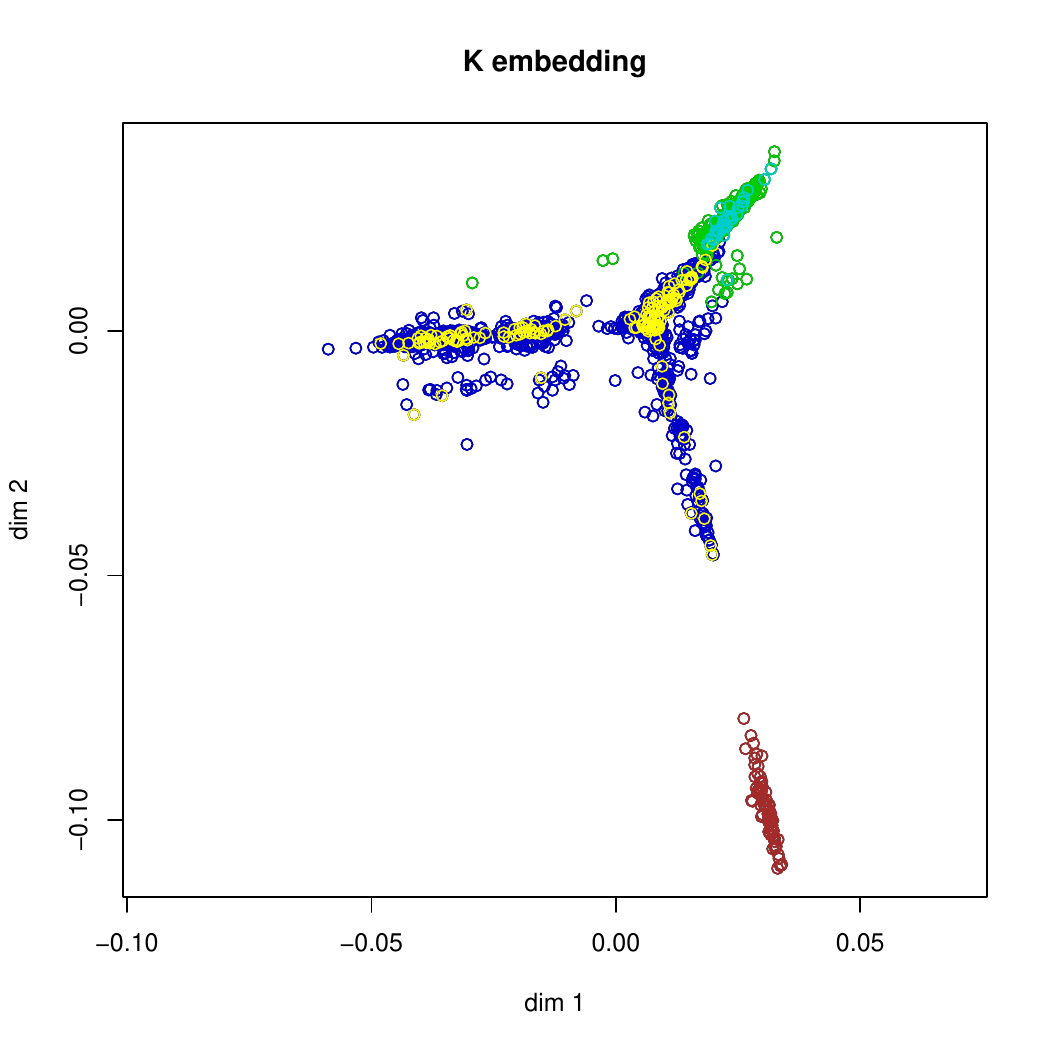} %
 \end{center}
\caption{A glance at the $K$-based clustering   for three different word embeddings.
Left: WikiGloVe, Center: TweetGloVe, Right: TVS.
  Different colors reflect the clusters. Dataset TWT.3.  
}\label{fig:TWT3ht_Kbased_cmp}
\end{figure}

Figure \ref{fig:TWT3ht_Kbased_cmp} shows similar behavior for TweetGloVe, while TVS provides a different picture - of several "thick lines". 
As Table \ref{tab:error_cmp} shows, $K$-based clustering under TVS performs worst.

\begin{figure}
\begin{center}
\includegraphics[width=0.3\textwidth]{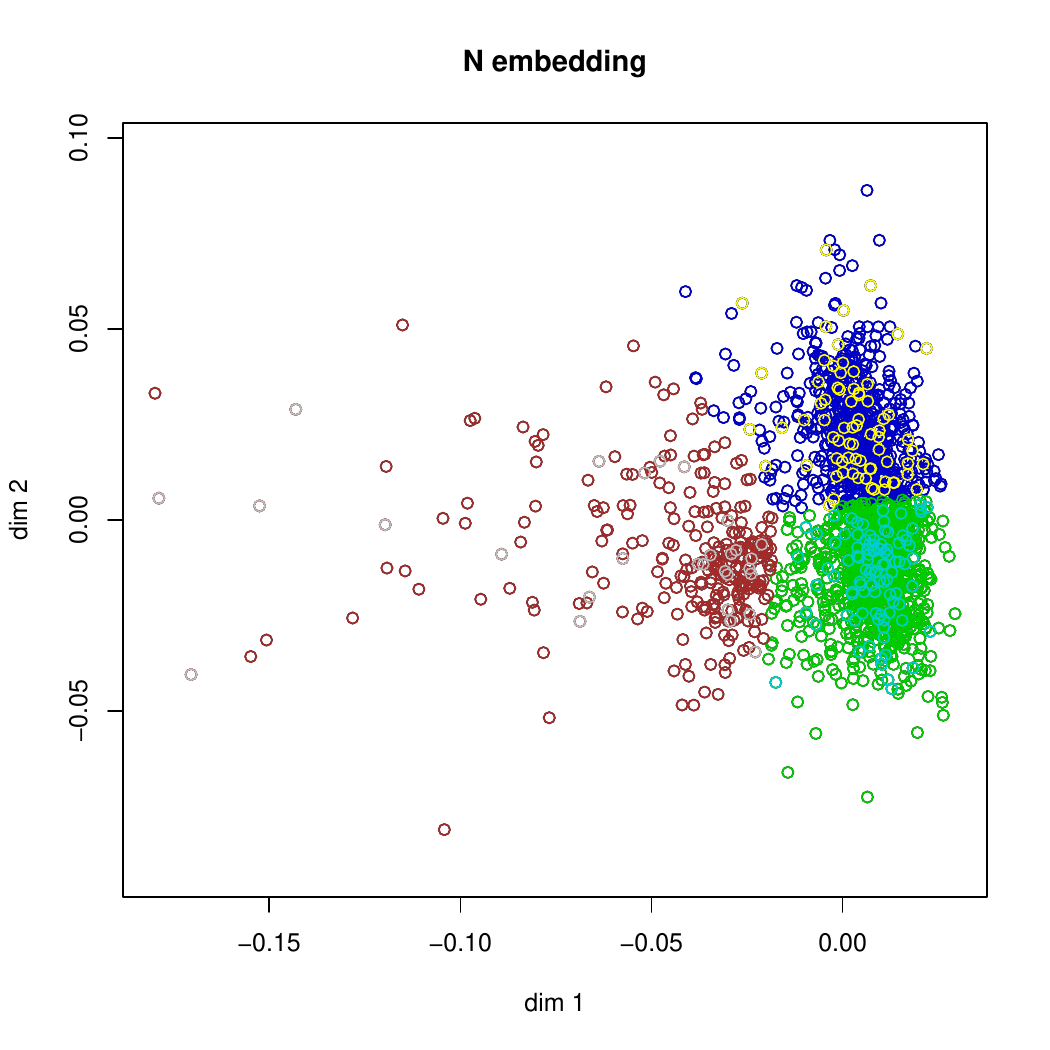} %
\includegraphics[width=0.3\textwidth]{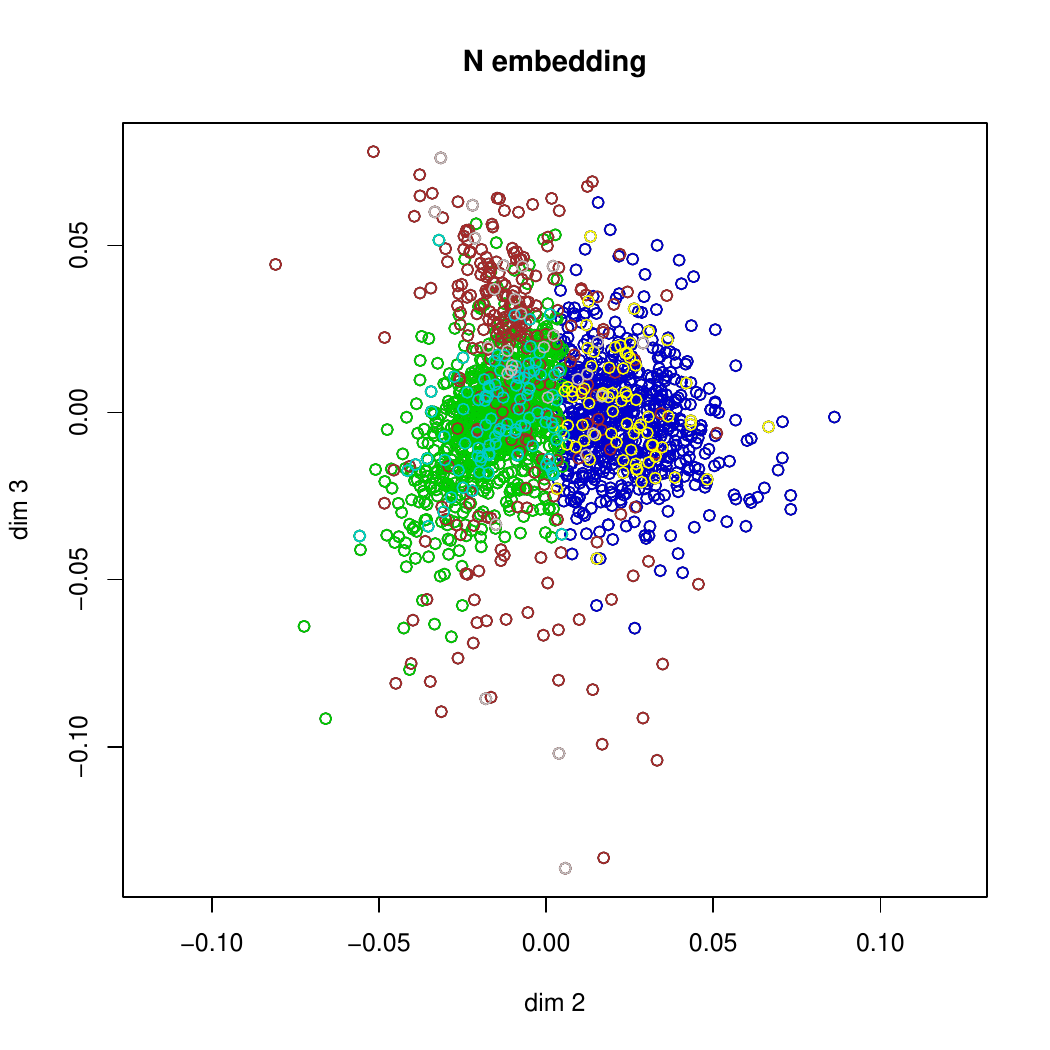} %
\includegraphics[width=0.3\textwidth]{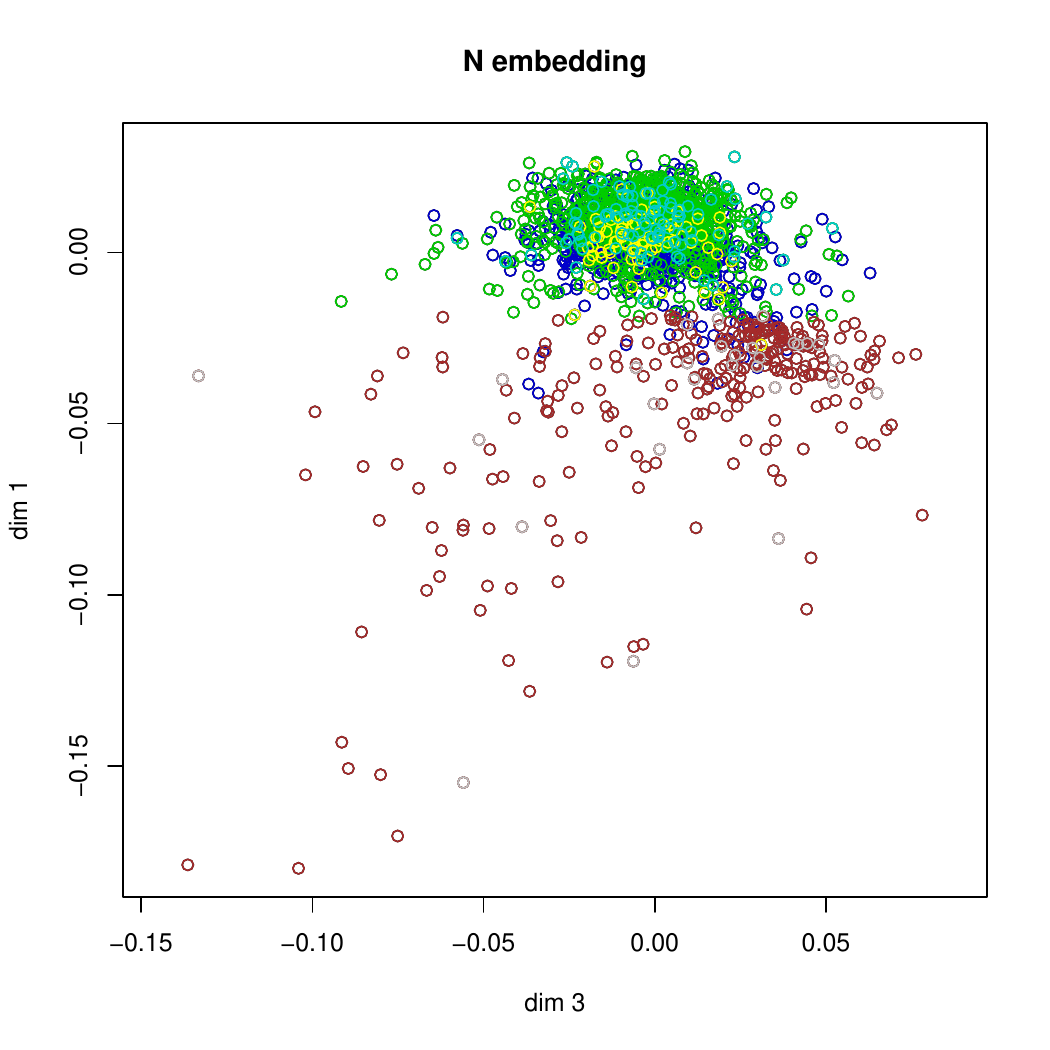} %
 \end{center}
\caption{A glance at the $N$-based clustering from three different perspectives (axes 1,2, or 2,3 or 1,3).  Different colors reflect the clusters.   Dataset TWT.3. Embedding WikiGloVe.
}\label{fig:TWT3ht_Nbased}
\end{figure}

Figure \ref{fig:TWT3ht_Nbased}
shows the clustering results obtained using the $N$-based method. Table \ref{tab:trueN} provides numerical information.  The documents appear to fill the space in a core region, while there are some documents scattered a bit away from it, but to a lesser extent than in the case of clustering based on $K$.   

\begin{figure}
\begin{center}
\includegraphics[width=0.3\textwidth]{TWT3ht_Nemb_12.pdf} %
\includegraphics[width=0.3\textwidth]{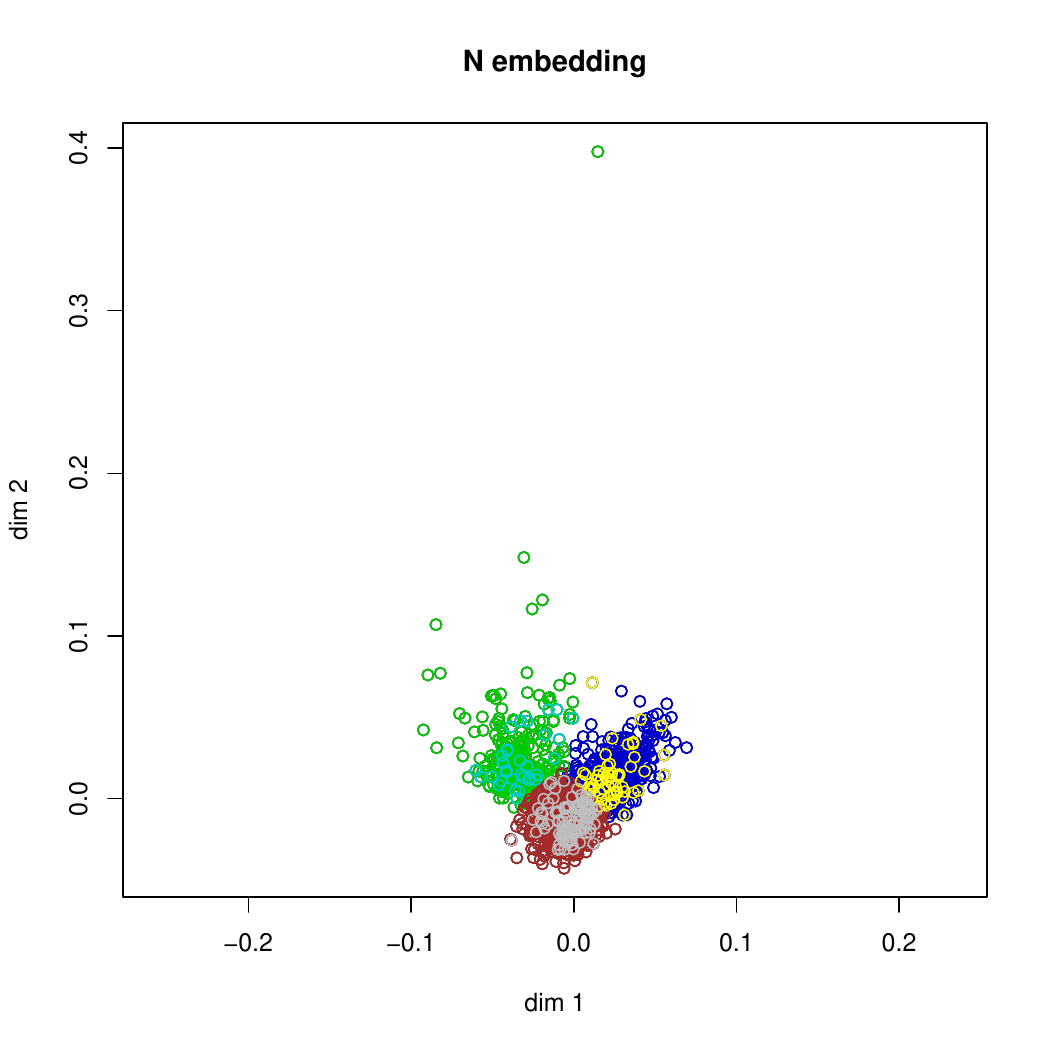} %
\includegraphics[width=0.3\textwidth]{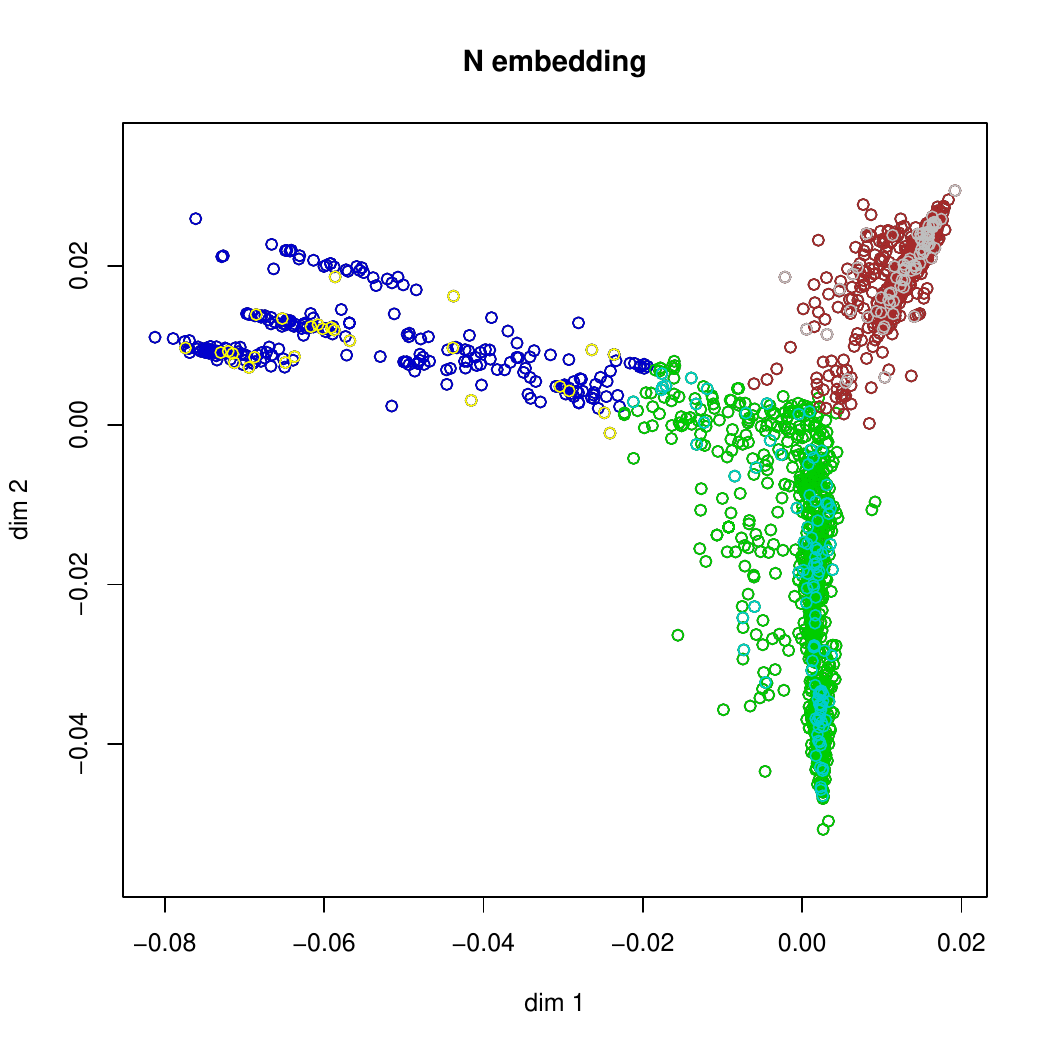} %
 \end{center}
\caption{A glance at the $N$-based clustering   for three different word embeddings.
Left: WikiGloVe, Center: TweetGloVe, Right: TVS.
  Different colors reflect the clusters.   Dataset TWT.3.  
}\label{fig:TWT3ht_Nbased_cmp}
\end{figure}

Table \ref{tab:error_cmp} shows slightly worse performance than $K$-based clustering for GloVe embedding (WikiGloVe, TweetGloVe). However, it is significantly better for TVS embedding. Figure \ref{fig:TWT3ht_Nbased_cmp} gives a visual impression similar to Figure \ref{fig:TWT3ht_Kbased_cmp}.

\begin{figure}
\begin{center}
\includegraphics[width=0.3\textwidth]{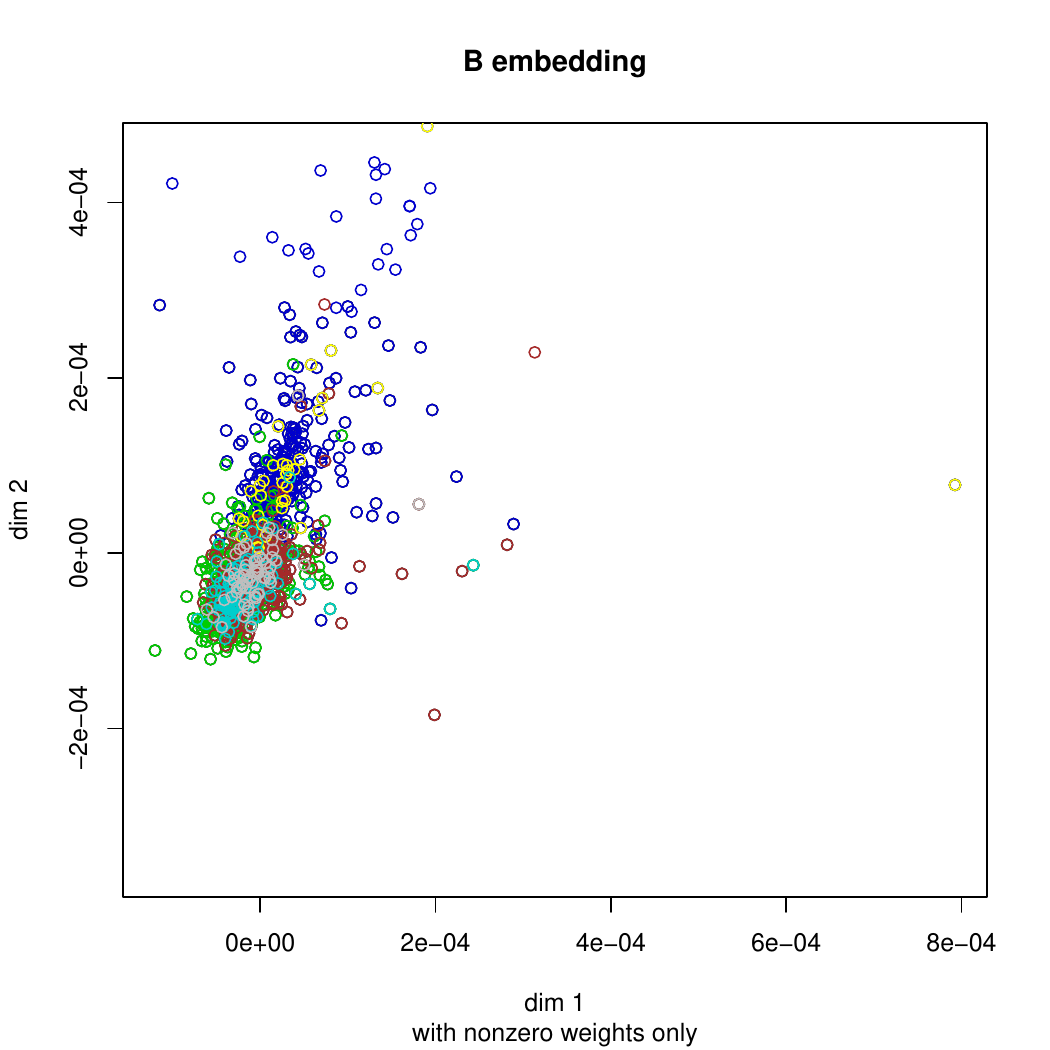} %
\includegraphics[width=0.3\textwidth]{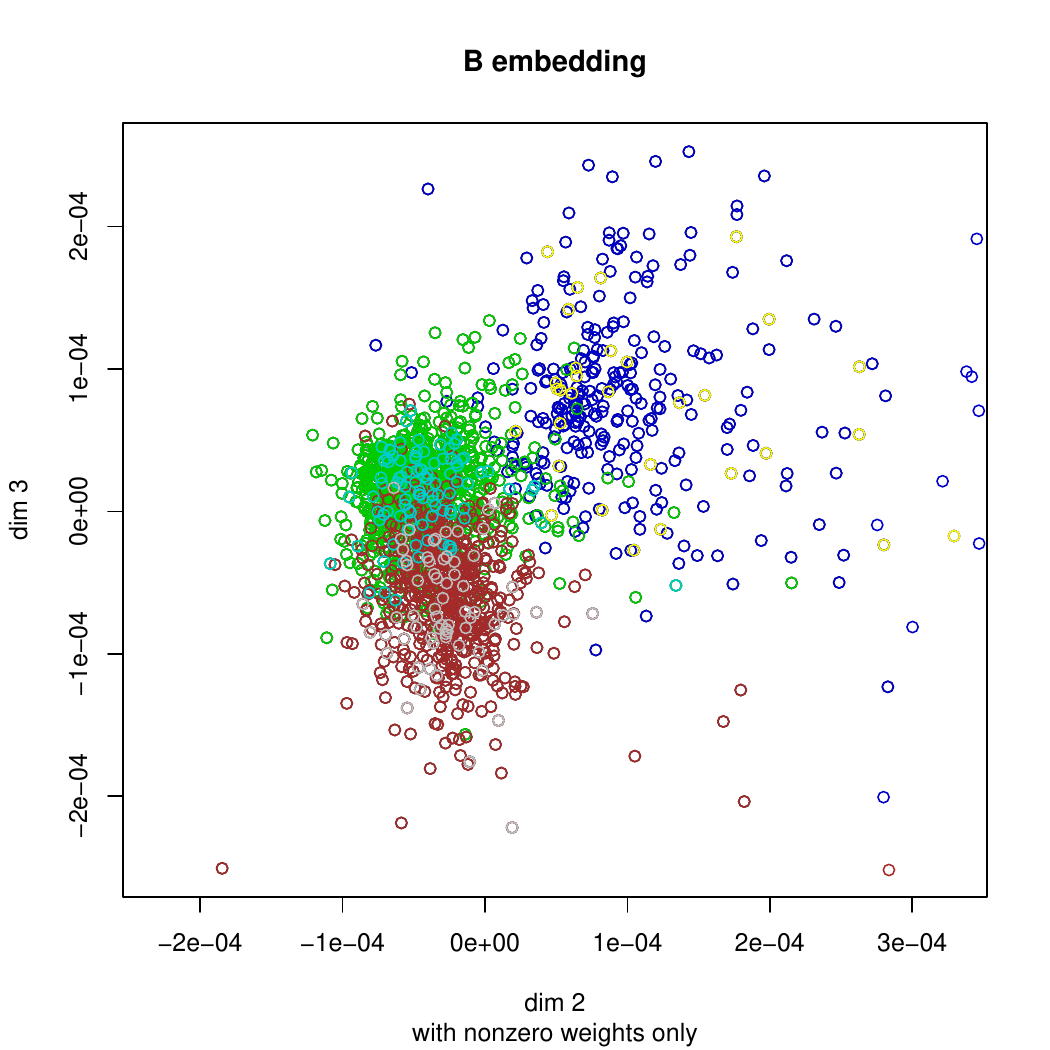} %
\includegraphics[width=0.3\textwidth]{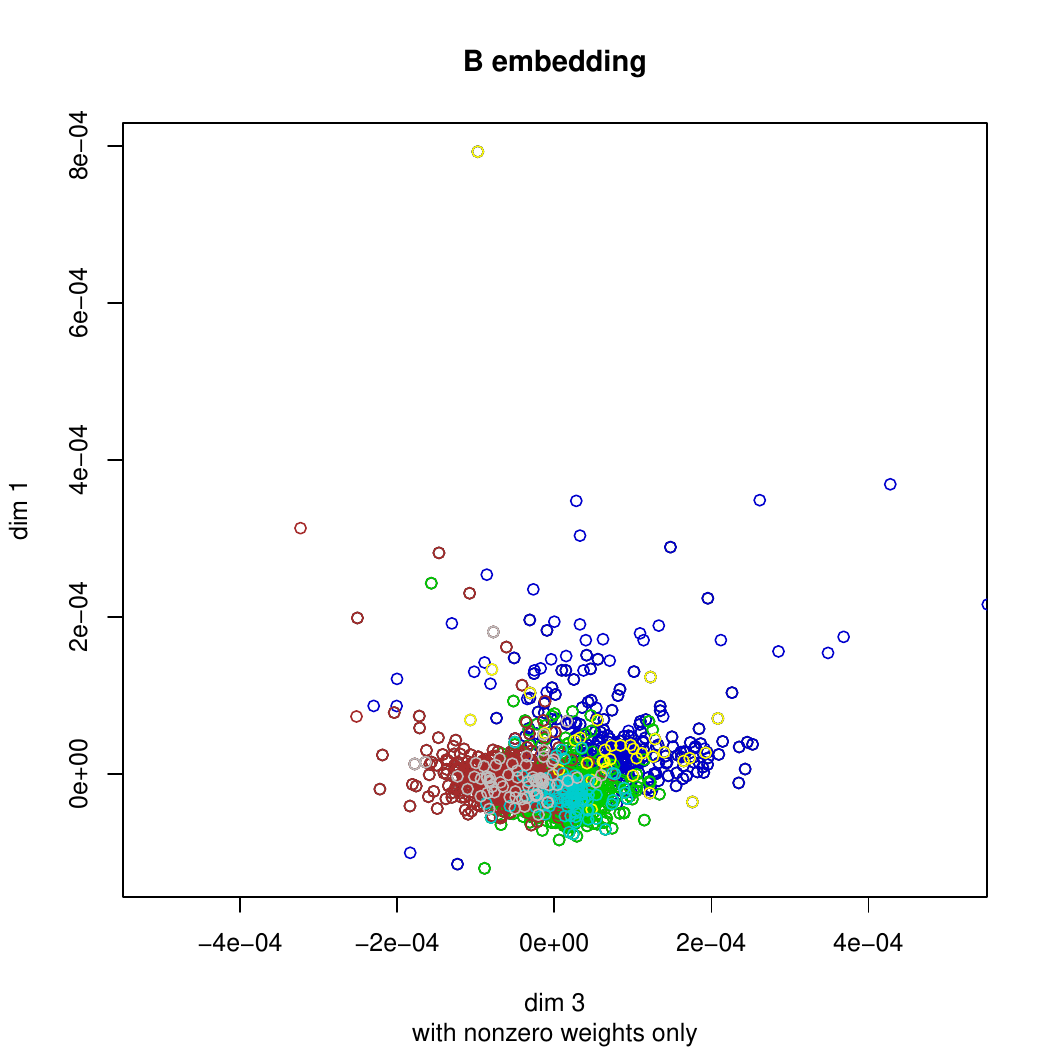} %
 \end{center}
\caption{A glance at the $\mathcal{B}$-based clustering from three different perspectives (axes 1,2, or 2,3 or 1,3).  Different colors reflect the clusters.   Dataset TWT.3. Embedding WikiGloVe.
}\label{fig:TWT3ht_Bbased}
\end{figure}

\begin{figure}
\begin{center}
\includegraphics[width=0.3\textwidth]{TWT3ht_Bemb_12.pdf} %
\includegraphics[width=0.3\textwidth]{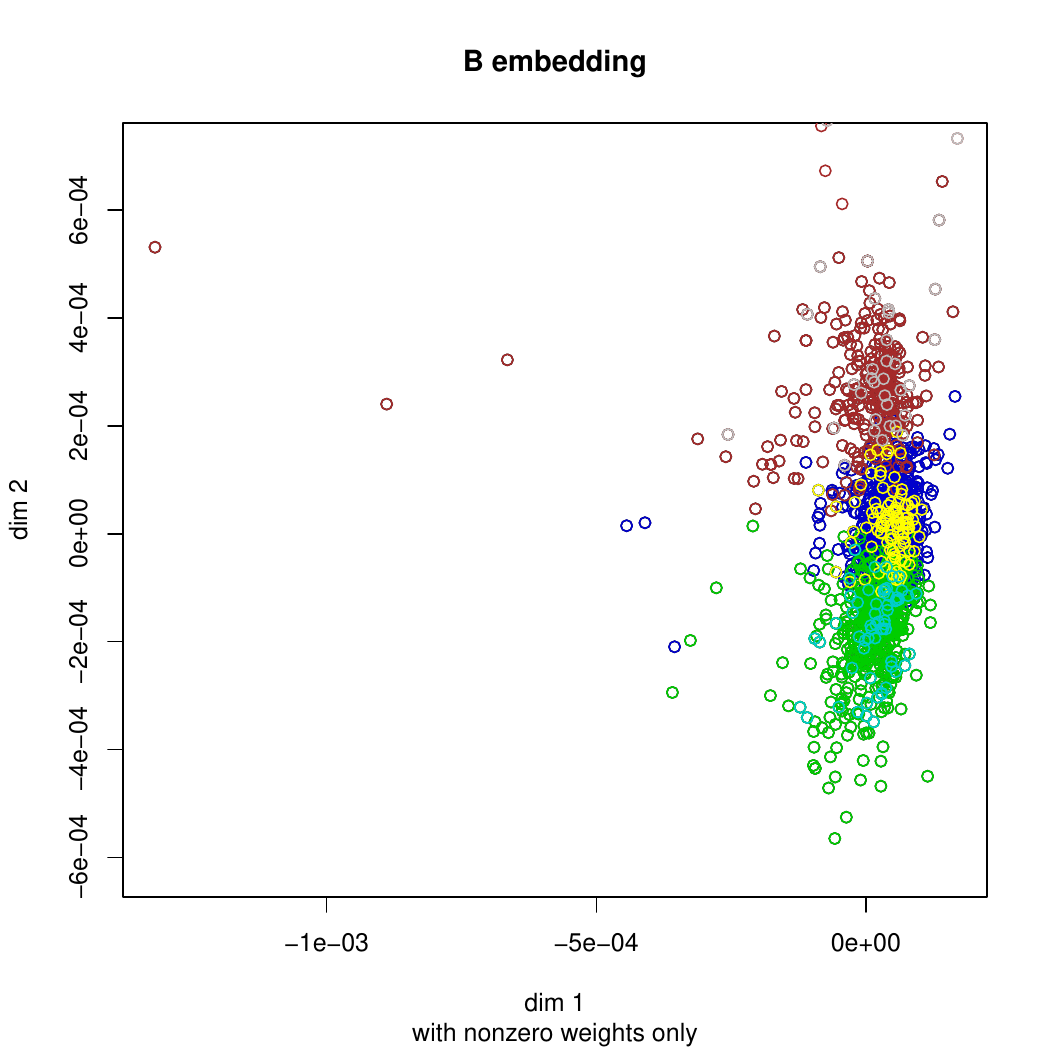} %
\includegraphics[width=0.3\textwidth]{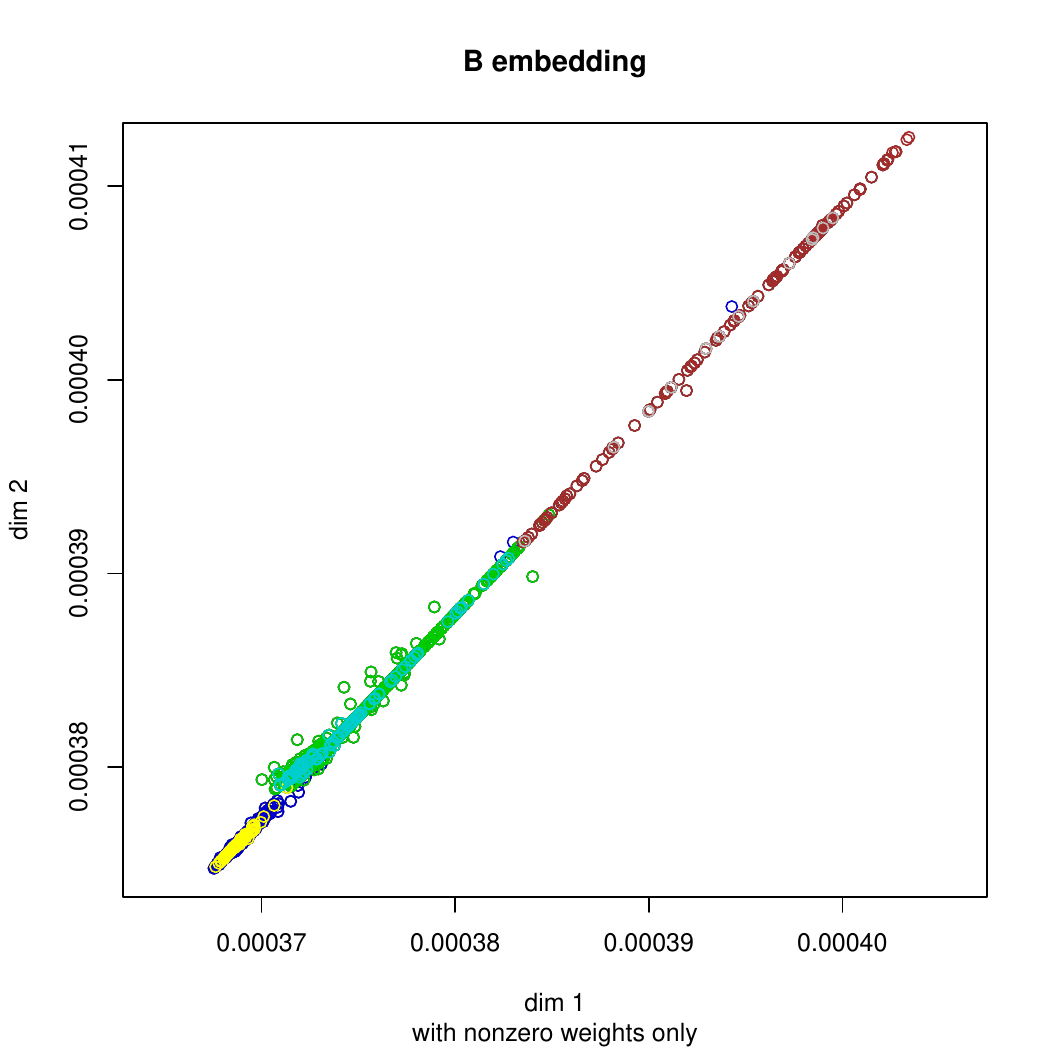} %
 \end{center}
\caption{A glance at the $\mathcal{B}$-based clustering for three different word embeddings.
Left: WikiGloVe, Center: TweetGloVe, Right: TVS.
  Different colors reflect the clusters.   Dataset TWT.3.  
}\label{fig:TWT3ht_Bbased_cmp}
\end{figure}

Figure \ref{fig:TWT3ht_Bbased}
shows the results of clustering using the $\mathcal{B}$-based method. Table \ref{tab:trueB} provides with numerical information.
As visible in Table \ref{tab:error_cmp}, it performs similarly to clustering based on $N$ in all embeddings. 
Figure \ref{fig:TWT3ht_Bbased_cmp} shows the flattening of the spacial structure of the embedding, with numerous datapoints away from a core area, while for the TVS embedding the flattening is extreme - to nearly a line. 

\begin{table} 
\centering
\begin{tabular}{|r|r|r|r|}
\hline TRUE/PRED& 1& 2& 3\\
\hline  \tht{anjisalvacion}& 0& 0& 732\\
\hline  \tht{nowplaying}& 1& 1& 437\\
\hline  \tht{puredoctrinesofchrist}& 0& 0& 831\\
\hline
\end{tabular}
\caption{Is true clustering implied with $L$-based GSC?
The number of elements in correct clusters 833, 
incorrectly clustered:  1169, 
 errors:  58.3\%. Dataset TWT.3. Embedding WikiGloVe.
 }
\label{tab:trueL}

\end{table} 


\begin{table} 
\centering
\begin{tabular}{|r|r|r|r|}
\hline TRUE/PRED& 1& 2& 3\\
\hline  \tht{anjisalvacion}& 707& 9& 16\\
\hline  \tht{nowplaying}& 142& 259& 38\\
\hline  \tht{puredoctrinesofchrist}& 176& 4& 651\\
\hline
\end{tabular}
\caption{Is true clustering implied with our $K$-based method? Number of elements in correct clusters 1617, 
incorrectly clustered:  385 
= errors:  19.2\% Dataset TWT.3. Embedding WikiGloVe.
 }
\label{tab:trueK}

\end{table} 

\begin{table} 
\centering
\begin{tabular}{|r|r|r|r|}
\hline TRUE/PRED& 1& 2& 3\\
\hline  \tht{anjisalvacion}&               692& 1  & 39\\
\hline  \tht{nowplaying}&                  90& 291 & 57\\
\hline  \tht{puredoctrinesofchrist}&       205& 0 & 626\\
\hline
\end{tabular}
\caption{Is true clustering implied with $N$-based GSC? Number of elements in correct clusters 1609, 
incorrectly clustered:  392, 
errors:  19.8\% Dataset TWT.3. Embedding WikiGloVe.
}
\label{tab:trueN}

\end{table} 


\begin{table} 
\centering
\begin{tabular}{|r|r|r|r|}
\hline TRUE/PRED& 1& 2& 3\\
\hline  \tht{anjisalvacion}&         679&  14& 39\\
\hline  \tht{nowplaying}&            74&   328& 36\\
\hline  \tht{puredoctrinesofchrist}&  238&   5&   588\\
\hline
\end{tabular}
\caption{Is true clustering implied with our $\mathcal{B}$-based method? Number of elements in correct clusters 1595, 
incorrectly clustered:  406, 
= errors:  20.3\% Dataset TWT.3. Embedding WikiGloVe.
}
\label{tab:trueB}
\end{table} 


\begin{table} 
\centering
\begin{tabular}{|r|r|r|r|r|}
\hline  & $L$-based& $K$-based& $N$-based &$\mathcal{B}$-based\\
\hline  WikiGloVe & 58.3\%  & 19.2\% & 19.8\%  & 20.3\%\\
\hline  TweetGloVe& 58.3\% &16.5\% & 20.22\% & 19.77\%\\
\hline  TVS& 58.4 \% & 42.0\% &  8.9\%  & 15.6\%\\
\hline
\end{tabular}
\caption{Error comparisons  
}
\label{tab:error_cmp}

\end{table} 

\FloatBarrier

\subsection{Explanations}
\label{sec:clusterExplain}

The tables below present explanations of class membership for various clustering methods for data set TWT.3 under WikiGloVe embedding.

Before turning to the results of the clustering methods, let us discuss the explanation results for the ``intrinsic'' clusters, that is, the clusters matching exactly the hashtags, as presented in tables 
\ref{tab:exp:True:WikiGloVe}, 
\ref{tab:exp:True:WikiGloVe:diff},
\ref{tab:exp:True:TwitGloVe}, 
\ref{tab:exp:True:TwitGloVe:diff},
\ref{tab:exp:True:fTVS}, 
\ref{tab:exp:True:fTVS:diff},

Table \ref{tab:exp:True:WikiGloVe} presents 50 words that are most similar to the gravity center of the cluster according to the formula
\eqref{eq:wordCluster:sim} under GloVe embedding. 
Table \ref{tab:exp:True:TwitGloVe} shows the same under the TweetGloVe embedding. 
Table \ref{tab:exp:True:fTVS} shows the same under the classical TVS embedding. 
Let us draw special attention to the cluster \tht{puredoctrinesofchrist} as it has self-explaining semantics. 
The comparison of these three tables shows that different words explain same cluster. The vast majority of these words make perfect sense, but it is somehow disturbing that the most significant words are "names" "security" "destroyed" in case of GloVe embedding, and "names" "worked" "cause"  in TweetGloVe embedding. On the other hand, the TVS has really reasonable leading terms: "king" "god" "proverbs". 

What can be the reason behind this behaviour? Note first that the TVS embedding uses all the words that occur in the documents, while the GloVe and TweetGloVe embeddings both have restricted, though big vocabularies (400,000 and 1,193,514 words/tokens, respectively) that change the set of words used in clustering (some 30\% of words get dropped at the very onset.)
In addition, the distribution of words in the embedding space differs. While in TVS, the distances between words (tf) are essentially the same between each pair of words, the words in GloVe spaces are not uniformly distributed. Also, the negative coordinates may play a role that cannot be easily foreseeable. 

For the aforementioned reasons, we examined word rankings in the context of distinguishing one cluster's gravity center from those of other clusters.
The respective settings are presented in tables 
 \ref{tab:exp:True:WikiGloVe:diff},  \ref{tab:exp:True:TwitGloVe:diff} and \ref{tab:exp:True:fTVS:diff}. 
The ranking of the words was computed according to the formula   
\eqref{eq:wordCluster:diff:sim} for respective embeddings. 
If we compare now Tables  \ref{tab:exp:True:WikiGloVe}
and  \ref{tab:exp:True:WikiGloVe:diff}, we see a significant semantic improvement for GloVe embedding.   
All fifty top words seem to come from the considered domain. 
If we compare now Tables  \ref{tab:exp:True:TwitGloVe}
and  \ref{tab:exp:True:TwitGloVe:diff}, an improvement is also visible, but this time the words typical for old-style Bibles pop up. This may be disappointing, as the TweetGloVe embedding was oriented toward tweet document corpora. 

\Bem{ ======  On the other hand, 
if we compare now Tables  \ref{tab:exp:True:fTVS}
and  \ref{tab:exp:True:fTVS:diff}, 
not much is gained (improvements are perceived in the second half of the list) and the results are not as good as for GloVe Table \ref{tab:exp:True:WikiGloVe:diff}. 
 ======== }
On the other hand, not much is gained by comparing Tables  \ref{tab:exp:True:fTVS} and  \ref{tab:exp:True:fTVS:diff}. Improvements in the second half of the list are noticeable, but the results are not as good as those for GloVe in Table  \ref{tab:exp:True:WikiGloVe:diff}.

So one may conclude that different embeddings have their strengths and weaknesses. 
 
Now, let us examine the explanations of clusters that were obtained during the clustering process.
Table  \ref{tab:exp:Kbased:WikiGloVe} shows the explanations for $K$-based clustering. 
Surprisingly, the third cluster in this table aligns more closely with the semantics of the \tht{puredoctrinesofchrist} concept than the intrinsic clustering in Table \ref{tab:exp:True:WikiGloVe}. 
The same can be said about the $N$-based (Table  \ref{tab:exp:Nbased:WikiGloVe}) and the $B$-based (Table  \ref{tab:exp:Bbased:WikiGloVe}) models.
Note that $L$-based clustering was not labeled with any explanation because the clustering itself is poor. 
The explanations for clusterings with differentiation seem to be plausible too for the differentiating explanations,
about $K$-based (Table   \ref{tab:exp:Kbased:WikiGloVe:diff}) 
 about $N$-based (Table   \ref{tab:exp:Nbased:WikiGloVe:diff}) and  $B$-based (Table   \ref{tab:exp:Bbased:WikiGloVe:diff}). 

\begin{table}
\caption{True clustering explaining words. Dataset TWT.3. Embedding WikiGloVe.}    
\label{tab:exp:True:WikiGloVe}
\begin{tabular}{|p{0.088\textwidth}|p{0.84\textwidth}|}
\hline
Cluster & Top 50 explaining words  \\
\hline
\vct{35}{\tht{anjisalvacion}}

 &  [1] "fun", "working", "expectations", "win", "problems", "celebrate", "probably", "front", "catch", [10] "check", "experience", "think", "percent", "event", "group", "difficult", "behind", "key", "question", [20] "really", "during", "clear", "nervous", "myself", "situation", "making", "worse", "reasons", "ten", [30] "difficulty", "guys", "trying", "read", "battle", "four", "ground", "anymore", "respond", "source", [40] "each", "changed", "anything", "relationship", "game", "fact", "upset", "something", "could", "enough", "clinton" \\
\hline
\end{tabular}

\end{table}

\begin{table}
\caption{K-based  clustering    explaining words. Dataset TWT.3. Embedding WikiGloVe.}    \label{tab:exp:Kbased:WikiGloVe}
\end{table}

\begin{table}
\caption{N-based clustering   explaining words. Dataset TWT.3. Embedding WikiGloVe.}     \label{tab:exp:Nbased:WikiGloVe}

\end{table}

\begin{table}
\caption{B-based clustering   explaining words. Dataset TWT.3. Embedding WikiGloVe.}     \label{tab:exp:Bbased:WikiGloVe}

\end{table}

\begin{table}
\caption{True clustering differentiating  explaining words}    
\label{tab:exp:True:WikiGloVe:diff}

\end{table}

\begin{table}
\caption{K-based clustering   explaining words  - cluster differentiation}    
 \label{tab:exp:Kbased:WikiGloVe:diff}

\end{table}

\begin{table}
\caption{N-based clustering   explaining words  - cluster differentiation}    
 \label{tab:exp:Nbased:WikiGloVe:diff}
\begin{tabular}{|p{0.088\textwidth}|p{0.84\textwidth}|}
\hline
\vct{35}{Cluster 1} &

[1] "remix", "singles", "wav", "nano", "dj", "airplay", "flash", "edit", "wap", [10] "remastered", "abba", "gees", "radio", "vip", "colour", "facebook", "featuring", "app", "playlist", [20] "fm", "dave", "italiano", "e", "download", "recording", "streaming", "com", "jazz", "disc", [30] "info", "audio", "android", "email", "rapper", "video", "h", "fabolous", "tv", "station", [40] "mee", "jingle", "production", "soundtracks", "tempo", "ska", "music", "id", "spotify", "bee", [50] "en"
\\
\hline
\end{tabular}

\end{table}

\begin{table}
\caption{B-based clustering   explaining words  - cluster differentiation}  
 \label{tab:exp:Bbased:WikiGloVe:diff}
\begin{tabular}{|p{0.088\textwidth}|p{0.84\textwidth}|}
\hline
\vct{35}{Cluster 1} &

[1] "finally", "losing", "watching", "rules", "maybe", "win", "difficulty", "nervous", "problems", [10] "feeling", "too", "difficult", "worse", "expectations", "deep", "situation", "fun", "working", "tough", [20] "extended", "seem", "lock", "think", "confidence", "really", "enjoy", "wake", "expect", "feel", [30] "johnson", "upset", "mean", "tend", "could", "experience", "help", "overcome", "trying", "robert", [40] "less", "security", "probably", "lose", "opportunity", "sees", "enough", "question", "myself", "percent", [50] "fears"
\\
\hline
\end{tabular}

\end{table}

\begin{table}
\caption{True clustering explaining words. TweetGloVe embedding.}  
\label{tab:exp:True:TwitGloVe}
\begin{tabular}{|p{0.088\textwidth}|p{0.84\textwidth}|}
\hline
Cluster & Top 50 explaining words  \\
\hline
\vct{35}{\tht{anjisalvacion}} &
[1] "fun", "celebrate", "probably", "check", "working", "problems", "catch", "win", "anymore", [10] "reasons", "myself", "read", "worse", "expectations", "think", "during", "really", "worrying", "behind", [20] "battle", "yes", "tomorrow", "entire", "experience", "relationship", "respond", "grows", "trying", "yourself", [30] "front", "changed", "thinking", "guy", "emotions", "upset", "smile", "moving", "nervous", "football", [40] "someone", "group", "anything", "reply", "key", "happiness", "choices", "making", "needs", "makes", [50] "matters"
\\
\hline
\end{tabular}

\end{table}

\begin{table}
\caption{True clustering   explaining words. TweetGloVe embedding - differentiating.}  
\label{tab:exp:True:TwitGloVe:diff}
\begin{tabular}{|p{0.088\textwidth}|p{0.84\textwidth}|}
\hline
Cluster & Top 50 explaining words  \\
\hline
\vct{35}{\tht{anjisalvacion}} &
 [1] "fun", "worrying", "probably", "problems", "relationship", "expectations", "nervous", "catch", "anymore", [10] "reasons", "worse", "celebrate", "upset", "smile", "myself", "choices", "thinking", "working", "emotions", [20] "happiness", "friendship", "think", "mistake", "someone", "yourself", "really", "win", "moving", "difficult", [30] "trying", "behind", "realize", "grows", "respond", "changed", "happens", "anything", "luck", "happen", [40] "expect", "miss", "difference", "change", "reply", "havent", "during", "attitude", "feels", "guy", [50] "seem" 
 \\
\hline
\end{tabular}

\end{table}

\FloatBarrier

\begin{table}
\caption{True clustering   explaining words   under TVS embedding}  
\label{tab:exp:True:fTVS}
\begin{tabular}{|p{0.088\textwidth}|p{0.84\textwidth}|}
\hline
Cluster & Top 50 explaining words  \\
\hline
\vct{35}{\tht{anjisalvacion}} &

[1] "anjisalvacion", "anji", "dalampasigan", "life", "happy", "feelstheconcert", "birthday", "people", "salvacion", [10] "mv", "love", "yourself", "th", "cheers", "never", "day", "things", "withanji", "go", [20] "always", "success", "best", "than", "only", "let", "asap", "think", "world", "cant", [30] "crazy", "everything", "kasingkasing", "sa", "change", "others", "every", "keep", "failure", "then", [40] "time", "youre", "see", "d", "better", "heart", "live", "way", "find", "good", [50] "need"
\\
\hline
\end{tabular}

\end{table}

\FloatBarrier

\begin{table}
\caption{True clustering explaining words. TVS embedding - differentiating.}  
\label{tab:exp:True:fTVS:diff}
\begin{tabular}{|p{0.088\textwidth}|p{0.84\textwidth}|}
\hline
Cluster & Top 50 explaining words  \\
\hline
\vct{35}{\tht{anjisalvacion}} &

[1] "anjisalvacion", "anji", "dalampasigan", "happy", "feelstheconcert", "life", "birthday", "salvacion", "mv", [10] "people", "yourself", "cheers", "th", "never", "withanji", "success", "always", "day", "things", [20] "asap", "best", "kasingkasing", "think", "sa", "(-) WDlet", "cant", "love", "crazy", "change", [30] "failure", "everything", "others", "only", "go", "d", "than", "youre", "world", "mediacon", [40] "happiness", "then", "something", "dahan", "feelsconcert", "see", "keep", "way", "successful", "beautiful", [50] "need"
\\
\hline
\end{tabular}
\end{table}

\FloatBarrier


\subsection{Clustering for Various GSC Variants}
\label{sec:clusterGSCvariants}

In this section, we provide an overview of the effectiveness of embeddings with various GSC methods. We use WikiGloVe and TweetGloVe embeddings and compare them with three variants of much simpler TVS embeddings (variants based on CountVectorizer implementation in \texttt{sklearn}, Term frequency, and Term Frequency/Invert Document Frequency based on TfidfVectorizer with parameter use\_idf set to false and true respectively). 

In the experiments, we use the TWT.10 dataset. Various methods of spectral k-means clustering (as underpinning clustering method for $K$-based GSC) are investigated. 

We have selected $K$-based GSC for several reasons, including the consistent explanations in the previous subsection.

Clustering experiments were performed with popular Python libraries: numpy \cite{NumPy:2020}, scipy \cite{SciPy:2020}, scikit-learn \cite{sklearnAPI:2013} and soy clustering \cite{soyclustering:020}, which is an implementation of spherical $k$-means \cite{SKmeans:2020:113288}.
More precisely, we used 

\begin{enumerate}
    \item \texttt{SpectralClustering} class from scikit-learn with two distinct  settings of the \texttt{affinity} parameter: \texttt{precomputed} (affinity from similarity matrix) and \texttt{nearest\_neighbors} (affinity from graph of nearest neighbors) - as a representative of the classical spectral clustering, and 

    \item \texttt{SphericalKMeans} class from soyclustering with the following combinations of (\texttt{init}, \texttt{sparsity}) parameter pairs (the mentioned 6 versions, short names given for reference):
           "k++.n": \texttt{('k-means++', None)},
           "k++.sc": \texttt{('k-means++', 'sculley')},
           "k++.md": \texttt{('k-means++', 'minimum\_df')}, 
           "sc.n": \texttt{('similar\_cut', None)},
           "sc.sc": \texttt{('similar\-\_cut', 'sculley')},
           "sc.md": \texttt{('similar\_cut', 'minimum\_df')},
           and 
    \item  $K$-based clustering (our implementation, exploiting spherical $k$-means. 
    The same combinations of parameter pairs (versions) were used as for \texttt{SphericalKMeans}  above. 
    \item 
    The number of leading eigenvectors from the $K$ matrix, that was used in the clustering, is indicated in the 'Dim Red' column. We experimented with reduction to  $r=10,20,3577$  eigenvectors, as well as with the full set of eigenvectors ($r= 7155$).
    \end{enumerate}

As follows from Table \ref{tab:massiveCmp}, 
the best results are achieved for TfIdf embeddings for all variants of the clustering algorithm. 
The Glove embedding trained on Wiki data seems to follow. 
The fact that the GloVe embedding trained on Twitter data does not perform well seems to be the result of the fact that it  contains a large amount of trash. 

Generally, Glove embeddings do not seem to constitute an improvement for clustering purposes over the TVS embeddings.  One may suspect that it is due to the sparseness of text in the tweets and/or the dictionary limitations, excluding numerous words used in the tweets. 

Dimensionality reduction, as expected, reduces the correctness of clustering, though reducing the dimensions by half isquite reasonable. 

 

\begin{table}[]
    \centering
    \caption{A comparative study, based on F measure, of the effectiveness of clustering within various embeddings, under usage of different dimensionality reduction(Dim Red column) using various clustering algorithms (Skm cnf column) within the framework of GSC. }
    \label{tab:massiveCmp}
    \begin{tabular}{|l|l||l|l|l|l|l|}
 \hline 

Dim     & Skm cnf &      Wkiki- &       Tweet-&  CountVect &    TfIdf TVS       &Tf TVS
\\
 Red    &   &   GloVe&  Glove&    TVS&   & 
 \\ \hline
  10 & $ k++.md $&\  0.2601& \textit{ 0.231 }&\  0.2739& \textbf{ 0.4738 }&\  0.2778
\\   10 & $ k++.n $&\  0.2622& \textit{ 0.2308 }&\  0.2791& \textbf{ 0.4819 }&\  0.277
\\   10 & $ k++.sc $&\  0.2671& \textit{ 0.2238 }&\  0.2775& \textbf{ 0.482 }&\  0.2783
\\   10 & $ sc.md $&\  0.2983&\  0.3587&\  0.281& \textbf{ 0.4684 }& \textit{ 0.278 }
\\   10 & $ sc.n $&\  0.2961&\  0.3681& \textit{ 0.2806 }& \textbf{ 0.4718 }&\  0.2815
\\   10 & $ sc.sc $&\  0.3137&\  0.3552& \textit{ 0.2808 }& \textbf{ 0.4715 }&\  0.2809
\\   20 & $ k++.md $&\  0.2587& \textit{ 0.2431 }&\  0.276& \textbf{ 0.5063 }&\  0.2808
\\   20 & $ k++.n $&\  0.2578& \textit{ 0.2412 }&\  0.2779& \textbf{ 0.5044 }&\  0.2791
\\   20 & $ k++.sc $&\  0.2616& \textit{ 0.2366 }&\  0.2745& \textbf{ 0.5042 }&\  0.2755
\\   20 & $ sc.md $&\  0.3106&\  0.3549&\  0.2838& \textbf{ 0.4902 }& \textit{ 0.2773 }
\\   20 & $ sc.n $&\  0.2978&\  0.362&\  0.2834& \textbf{ 0.4986 }& \textit{ 0.2816 }
\\   20 & $ sc.sc $&\  0.3021&\  0.3594& \textit{ 0.2769 }& \textbf{ 0.493 }&\  0.2804
\\   3577 & $ k++.md $&\  0.2599& \textit{ 0.2383 }&\  0.2764& \textbf{ 0.5142 }&\  0.2778
\\   3577 & $ k++.n $&\  0.2583& \textit{ 0.2286 }&\  0.2809& \textbf{ 0.5209 }&\  0.2815
\\   3577 & $ k++.sc $&\  0.2587& \textit{ 0.2233 }&\  0.2834& \textbf{ 0.5052 }&\  0.2804
\\   3577 & $ sc.md $&\  0.3119&\  0.3705& \textit{ 0.2816 }& \textbf{ 0.5056 }&\  0.2845
\\   3577 & $ sc.n $&\  0.3084&\  0.3573&\  0.282& \textbf{ 0.4947 }& \textit{ 0.2808 }
\\   3577 & $ sc.sc $&\  0.299&\  0.3767& \textit{ 0.2784 }& \textbf{ 0.5006 }&\  0.2792
\\   None & $ k++.md $&\  0.4306&\  0.49&\  0.4082& \textbf{ 0.6648 }& \textit{ 0.405 }
\\   None & $ k++.n $&\  0.4173&\  0.4922&\  0.396& \textbf{ 0.6418 }& \textit{ 0.3902 }
\\   None & $ k++.sc $&\  0.413&\  0.5002& \textit{ 0.3877 }& \textbf{ 0.6408 }&\  0.3928
\\   None & $ sc.md $&\  0.4216&\  0.4749& \textit{ 0.3812 }& \textbf{ 0.6763 }&\  0.4045
\\   None & $ sc.n $&\  0.4107&\  0.488& \textit{ 0.3977 }& \textbf{ 0.653 }&\  0.404
\\   None & $ sc.sc $&\  0.4221&\  0.4815& \textit{ 0.3913 }& \textbf{ 0.6383 }&\  0.3922
\\      \hline

    \end{tabular}
\end{table}

\FloatBarrier



\section{Future Work - Other embeddings, including non-linear ones }
\label{sec:future}

In this paper, we extend the idea of explainability of Graph Spectral Clustering results for textual documents to GloVe embedding, given that document similarity is computed using cosine similarity, which was originally designed for term vector space \cite{Plosone2025}. 
In this way, information about word relationships, inherent for GloVe embedding, was combined with the document graph relationship, inherent for Graph Spectral Clustering embedding. 

To extend explainability from term vector space embeddings to GloVe embeddings, the complexities resulting from nonorthogonalities of word embeddings in GloVe had to be handled in a way that allows for efficient computations.  

The experiments performed show that term vector space embeddings are
more advantageous than GloVe embeddings for short documents like Twitter posts, if the standard GSC clustering methods are used (L-based, N-based), though nonstandard K-based GSC benefits from GloVe. 
However,  we demonstrate via an example that an explanation built in the GloVe embedding may turn out to be more appealing than in term vector space embedding. This may be considered an important advantage because of the pressure for explainability of results of clustering in practical settings. Possibly, a mixture of both TVS and GloVe embeddings may benefit the explanations. Future research is needed.

In our experiments, we observed the problem that for some document sets the GloVe embedding causes problems for GSC because similarities may turn out to be negative. We propose a method for handling this issue in a separate paper
\cite{MAK:STW:BST:DCZ:PBR:2025:handlingnegativesimilarities}

While our method of explanation relies on the linearity of  word-document-embedding transformation and is extensible to this category of embeddings, we are aware that 
there have also been proposed non-linear transformations between document embedding and word embedding, the best known being   
Doc2Vec \cite{lau2016empiricaldoc2vec},   and BERT \cite{devlin2019bert}, 
and that there exist many other approaches too.    

\cite{Ye:2016} suggests an implicit embedding by assuming that the similarity between documents shall be seen as accordingly weighted similarity of most similar words in both documents. 
Same approach, though a bit different in details, is presented by \cite{Farouk:2020}.

\cite{vor-der-bruck:2019} suggests to compute the matrix of similarities of all words in two documents and then to use a matrix norm to calculate the similarity between documents. 
Under some norms (L11 norm if the word similarities cannot be negative under an embedding) this can be reduced to the aforementioned linear combination of words as an embedding of a document. 

\cite{Kenter:2015:STS:2806416.2806475} makes use of a complex weighted transformation of embedding of words to get similarity of documents which makes the relationship to document embedding non-linear. 

\cite{De_Boom:2016} proposes a non-linear embedding method particularly devoted to short texts, combining frequency and semantic information. Starting with  idf's, as word embeddings a loss function comparing each pair of documents is optimized to obtain final document embeddings.  

These and other methods pose a new challenge for explaining document membership in clusters based on these embeddings, as the impact of a word on the cluster is not that simple as in linear embeddings, as studied in this paper. The challenges of non-linearity are subject of our on-going research. 

\bibliographystyle{plain}


\end{document}